\documentclass{article}

\usepackage[final]{corl_2026}
\makeatletter
\renewcommand{\@conferencelocation}{Austin, Texas, USA}
\makeatother
\usepackage{graphicx}
\usepackage{amsmath}
\usepackage{amsfonts}
\usepackage{siunitx}
\usepackage{booktabs}
\usepackage{multirow}
\usepackage{tabularx}
\usepackage{array}
\newcolumntype{Y}{>{\centering\arraybackslash}X}
\usepackage{hyperref}
\usepackage{cleveref}
\newcommand{\refapp}[1]{\hyperref[#1]{App.~\ref*{#1}}}
\usepackage{wrapfig}
\usepackage{float}

\title{CoopUQ: Robust Multi-Robot Social Navigation via Uncertainty-Aware Cooperative Forecasting}

\hypersetup{
  pdftitle={CoopUQ: Robust Multi-Robot Social Navigation via Uncertainty-Aware Cooperative Forecasting},
  pdfauthor={Zehao Wang, Jianpeng Yao, Yuping Wang, Yiping Li, Jiaqi Li, Zhixu Li, Kanghoon Lee, Jinfeng Wang, Cong Liu, Zhaowei Tan, Jinkyoo Park, Jiachen Li}
}

\author{
  \textbf{Zehao Wang\textsuperscript{1} \quad Jianpeng Yao\textsuperscript{2} \quad Yuping Wang\textsuperscript{3} \quad Yiping Li\textsuperscript{4}}\\
  \textbf{Jiaqi Li\textsuperscript{5} \quad Zhixu Li\textsuperscript{2} \quad Kanghoon Lee\textsuperscript{6} \quad Jinfeng Wang\textsuperscript{2}}\\
  \textbf{Cong Liu\textsuperscript{2} \quad Zhaowei Tan\textsuperscript{2} \quad Jinkyoo Park\textsuperscript{6} \quad Jiachen Li\textsuperscript{1}\begin{NoHyper}\thanks{Corresponding author: jiachen\_li@gatech.edu}\end{NoHyper}}\\[0.5em]
  \normalfont\textsuperscript{1}Georgia Institute of Technology \quad \textsuperscript{2}University of California, Riverside\\
  \normalfont\textsuperscript{3}University of Michigan, Ann Arbor \quad \textsuperscript{4}University of Georgia\\
  \normalfont\textsuperscript{5}New York University \quad \textsuperscript{6}KAIST
}

\begin{document}
\maketitle

\begin{abstract}
Reinforcement learning enables mobile robots to make fast and scalable navigation decisions in constrained human-shared environments. However, directly extending single-robot approaches to multi-robot settings often fails to ensure safety and efficiency, especially under occlusions and out-of-distribution (OOD) scenarios, where individual robots may overlook potential risks or make inaccurate predictions. Leveraging cooperation effectively for both prediction and navigation remains a major challenge. We propose CoopUQ, a multi-robot social navigation framework that bridges uncertainty-aware cooperative occupancy forecasting and robust decision making. CoopUQ predicts future occupancy grid maps through cross-robot spatiotemporal feature fusion and further leverages forecasts and their uncertainty to guide policy via constrained multi-agent reinforcement learning, where the cost penalizes navigation through regions predicted to be occupied or highly uncertain. This design enables the policy to adapt its risk sensitivity under distribution shifts. Experimental results across multiple Unity3D-based simulation environments with both in-distribution and OOD settings, including diverse layouts, pedestrian densities, and rushing behavior, demonstrate improved prediction performance and safety over representative baselines. Real-robot experiments further demonstrate its robustness across various scenes. Our code and videos are available at \href{https://coopuq.github.io/}{https://coopuq.github.io/}.
\end{abstract}


\section{Introduction}
Safely navigating among humans is fundamental to deploying mobile robots in real-world shared spaces such as hospitals, offices, and warehouses~\citep{kruse2013human, rios2015proxemics, mavrogiannis2023core, chen2020relational}. Multiple robots may operate simultaneously in these constrained spaces while pedestrians move unpredictably through narrow corridors, blind corners, and cluttered layouts. Navigation decisions are shaped by both static geometry and dynamic human motion. Reinforcement learning (RL) and multi-agent reinforcement learning (MARL) support reward-driven policy learning and preference-based refinement~\citep{zhao2026evonav,kim2025human}, provide fast decision-making, and have shown promising performance in multi-robot socially-aware navigation~\citep{chen2017decentralized, han2020cooperative, long2018towards, wang2024multi, chen2017socially, everett2018motion, chen2019crowd, perez2021robot, xie2023drl}.
Previous studies have shown that prediction-aware policies can help robots shape their decisions on future human motion~\citep{liu2023intention, li2024multi, bruckshen2020human, wang2022group, poddar2023crowd}. While trajectory prediction is widely used to represent pedestrian futures, it often focuses on dynamic agents alone and does not model the surrounding static geometry that is crucial in constrained spaces. Occupancy grid maps (OGMs)~\citep{schreiber2022multitask, xie2025scope} provide a more unified representation, encoding static obstacles, free space, and human motion within the same spatial structure. This makes OGMs suitable for social navigation in cluttered and partially observable environments.

\begin{figure}[!t]
    \centering
    \includegraphics[width=\linewidth]{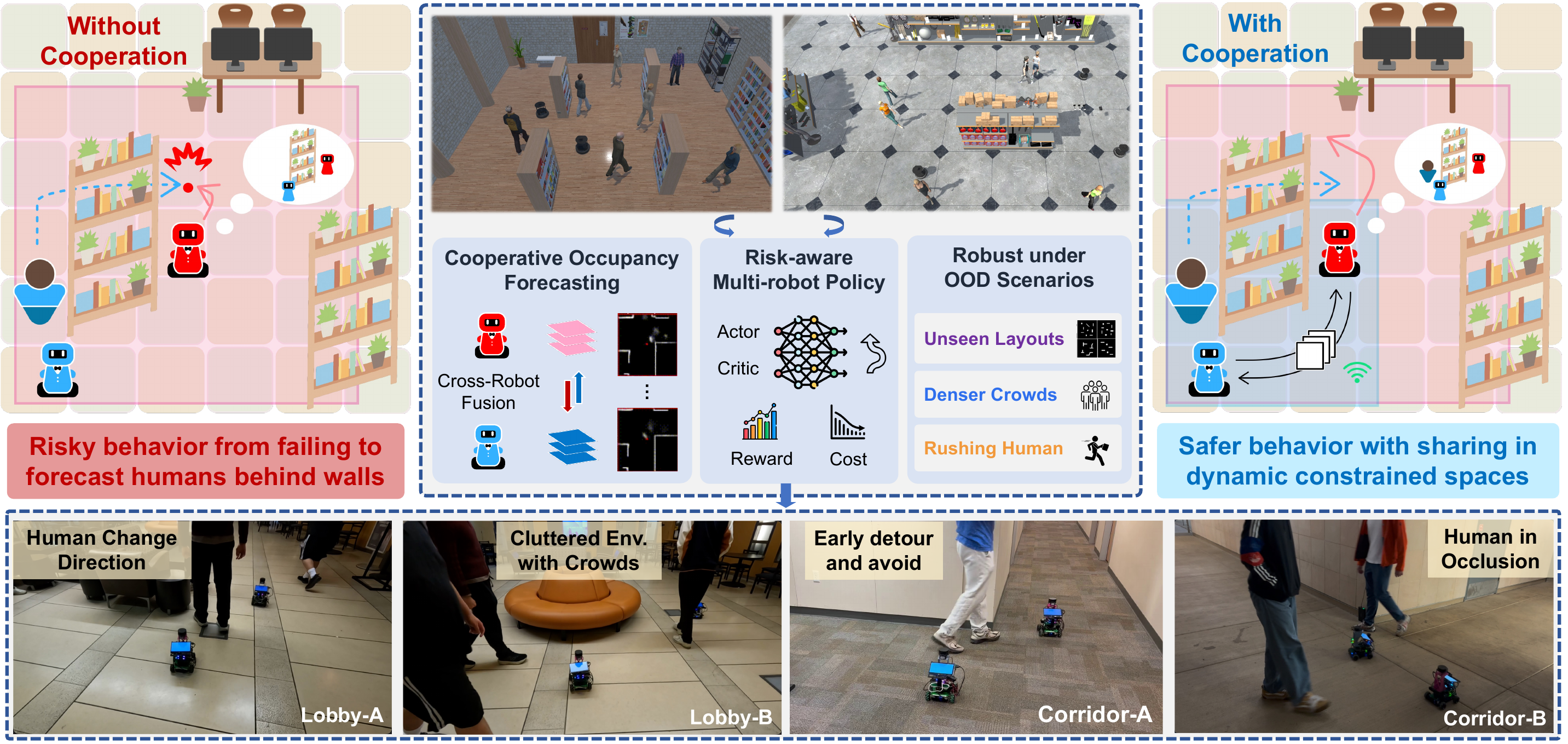}
    \caption{The single-agent approach relies only on the robot's own observations and misses an incoming pedestrian behind the obstacle. In contrast, CoopUQ fuses shared information, anticipates dynamic motions under uncertainty, and plans a safe trajectory. Closed-loop experiments in Unity-3D simulation demonstrate robustness in OOD scenarios, and the learned policy transfers directly to various real-world environments, achieving safe navigation across diverse crowds and layouts.}
    \vspace{-0.5cm}
\label{fig:motivation}
\end{figure}

However, forecasting future occupancy remains challenging in environments with evolving pedestrian intentions, diverse motion patterns, denser crowds, and unseen occlusion-heavy regions. Learning-based policies often degrade under such out-of-distribution (OOD) scenarios, since distribution shifts can introduce prediction errors and severely mislead planning. This challenge is pronounced when multiple robots operate in constrained environments shared with pedestrians, where each individual robot may overlook risks caused by occlusions or inaccurate predictions. Therefore, effective cooperation among robots is essential to ensure safety, efficiency, and human comfort. 
Moreover, to ensure robust and adaptive navigation, robots need not only to predict future states but also to estimate the uncertainty of predictions. Such uncertainty indicates when a prediction is reliable and when the policy should adapt its behavior.
For instance, as shown in Figure~\ref{fig:motivation}, a robot may get stuck in a confined area due to static obstacles or fail to detect pedestrians hidden behind a wall when turning a corner. In contrast, cooperative multi-robot systems can coordinate planning and exchange information to improve task execution and situational awareness~\citep{gupta2026scaleplan,lee2026priority,zhang2026commcp}. By sharing spatiotemporal features when a nearby robot detects a pedestrian in an occluded region, robots can make more informed and proactive decisions, improving safety, efficiency, and human comfort under challenging distribution shifts.

To this end, we propose \textbf{CoopUQ}, a multi-robot social navigation framework. CoopUQ forecasts future OGMs and estimates uncertainty maps by modeling the stochasticity of future occupancy over cross-robot spatiotemporal features, improving prediction quality over raw-LiDAR or prediction-level fusion. Beyond using uncertainty as an extra observation, we employ constrained multi-agent reinforcement learning (CMARL)~\citep{gu2023safe} to explicitly guide risk-aware control, allowing robots to avoid ambiguous regions under distribution shifts while maintaining efficiency in well-observed areas.
Our main contributions are three-fold. First, CoopUQ leverages uncertainty as a bridge between stochastic cooperative forecasting and robust decision making. Second, we design a deviation-guided cross-robot fusion module that adaptively aggregates complementary features in regions affected by occlusions or cross-robot disagreement. Third, we introduce a risk-aware safety cost in CMARL to explicitly guide the navigation policy. Extensive Unity3D simulations and real-robot experiments demonstrate safer and more robust navigation across in-distribution and OOD scenarios.

\section{Related Work}

\noindent \textbf{Social Robot Navigation.}
Mobile robots that navigate in spaces shared with humans should adhere to social norms \cite{mirsky2024conflict, chen2020relational, zhu2024learn, van2008reciprocal, chandra2025deadlock, pouria2024topology}. Social navigation is often modeled as a Partially Observable Markov Decision Process, where agents act based on incomplete local observations \cite{mirsky2024conflict}.
A key challenge is modeling partial and uncertain observations in real-world environments~\citep{jiang2026navtrust}.
Many prior works focus on predicting trajectories of dynamic obstacles, modeling pedestrians as simple position-radius tuples. These rule-based or learning-based methods perform well in simulation \cite{liu2023intention, li2024multi, wang2024multi, chen2017decentralized, han2020cooperative, long2018towards, escudie2024attention}, but often assume full visibility in open spaces and overlook static obstacles, limiting their applicability in occluded or cluttered environments.
In contrast, we adopt OGM forecasting with uncertainties as a holistic spatiotemporal representation and aggregate complementary observations through multi-robot cooperation, enabling safer decisions in constrained environments.

\noindent \textbf{Occupancy Grid Map (OGM) Forecasting.}
Predicting the future environment remains challenging due to its inherent stochasticity and uncertainty. OGM is an effective spatial representation that unifies both static and dynamic obstacles.
Since OGMs can be treated as images, multi-step forecasting is often formulated as a video prediction problem \cite{mohajerin2019multi, hoermann2018dynamic, toyungyernsub2022dynamics, lange2025self, toyungyernsub2024predicting}. Recent studies use OGMs to model dynamic human crowds \cite{dugas2021navrep,xie2023stochastic}.
However, individual OGM forecasting remains challenging in constrained environments, where occlusions can lead to incomplete or less accurate predictions.
In contrast, cooperative forecasting allows multiple robots to fuse spatiotemporal features and produce uncertainty-aware predictions, constructing more reliable guidance for downstream navigation.

\noindent \textbf{Cooperative Perception and Prediction.}
Cooperative perception and prediction are crucial for addressing partial observability and perceptual limitations, particularly in environments with occlusions~\citep{lee2026cooperative}.
By sharing sensor data, intermediate features, or predictions, multiple agents can construct a more comprehensive scene understanding, enhancing detection and motion prediction in occlusion scenarios~\cite {xu2023cobevt,wang2025cmp,gao2025stamp,li2025comamba}.
In this work, we bridge cooperative occupancy forecasting with uncertainty-aware planning for multi-robot social navigation, allowing robots to leverage predictions and their corresponding uncertainties to make more robust decisions under occlusion and distribution shifts.

\section{Method}
\label{sec:method}

\subsection{Problem Formulation}

We consider the problem of navigating $N$ mobile robots in an environment with static obstacles (e.g., walls, bookshelves) and dynamic obstacles (e.g., pedestrians, other robots). Each robot is assigned a goal location, and our objective is to learn a multi-robot navigation policy that enables all robots to reach their goals safely and efficiently.
Formally, we formulate multi-robot social navigation as a constrained Markov game~\cite{gu2023safe}, defined by the tuple 
$\langle \mathcal{I}, \mathcal{S}, \boldsymbol{\mathcal{O}}, \boldsymbol{\mathcal{A}}, \mathcal{T}, R, C, \rho_0, \gamma \rangle$.
The agent set $\mathcal{I}=\{1,\ldots,N\}$ represents the robots. The state space $\mathcal{S}$ includes states of robots, dynamic, and static obstacles. The joint observation space is $\boldsymbol{\mathcal{O}}=\times_{i\in\mathcal{I}}\mathcal{O}^i$, where $\mathcal{O}^i$ is the local observation space of robot $i$. 
The joint action space is $\boldsymbol{\mathcal{A}}=\times_{i\in\mathcal{I}}\mathcal{A}^i$, where each action $a_t^i\in\mathbb{R}^2$ represents the desired 2D velocity of robot $i$ at time $t$. The transition function $\mathcal{T}$ governs the state evolution, with each robot tracking its desired velocity through a controller. 
$R$ and $C$ denote the reward and cost functions, which aggregate individual rewards and costs, respectively. Finally, $\rho_0$ is the initial state distribution and $\gamma\in[0,1)$ is the discount factor.
We optimize a shared policy $\pi_\theta$ that maximizes the expected cumulative reward while keeping the cumulative cost below a specified limit $\epsilon$:
\begin{equation}
    \begin{aligned}
    \max_{\pi_\theta} \quad 
     J^R(\pi_\theta)
    =
    \mathbb{E}_{\pi_\theta}
    \left[
    \sum_{t=0}^{T}
    \gamma^t R(s_t,\mathbf{a}_t)
    \right], \quad
    \text{s.t.} \quad 
     J^C(\pi_\theta)
    =
    \mathbb{E}_{\pi_\theta}
    \left[
    \sum_{t=0}^{T}
    \gamma^t C(s_t,\mathbf{a}_t)
    \right]
    \leq \epsilon,
    \end{aligned}
    \label{eq:constrained_opt}
\end{equation}
where $\mathbf{a}_t=(a_t^1,\ldots,a_t^N)$, $a_t^i \sim \pi_{\theta}(\cdot \mid \mathbf{o}_t^i)$, and $\mathbf{o}_t^i \in \mathcal{O}^i$ denotes the observation of robot $i$ at time $t$.

\subsection{Method Overview}
\begin{figure}[!t]
    \centering
    \includegraphics[width=\linewidth]{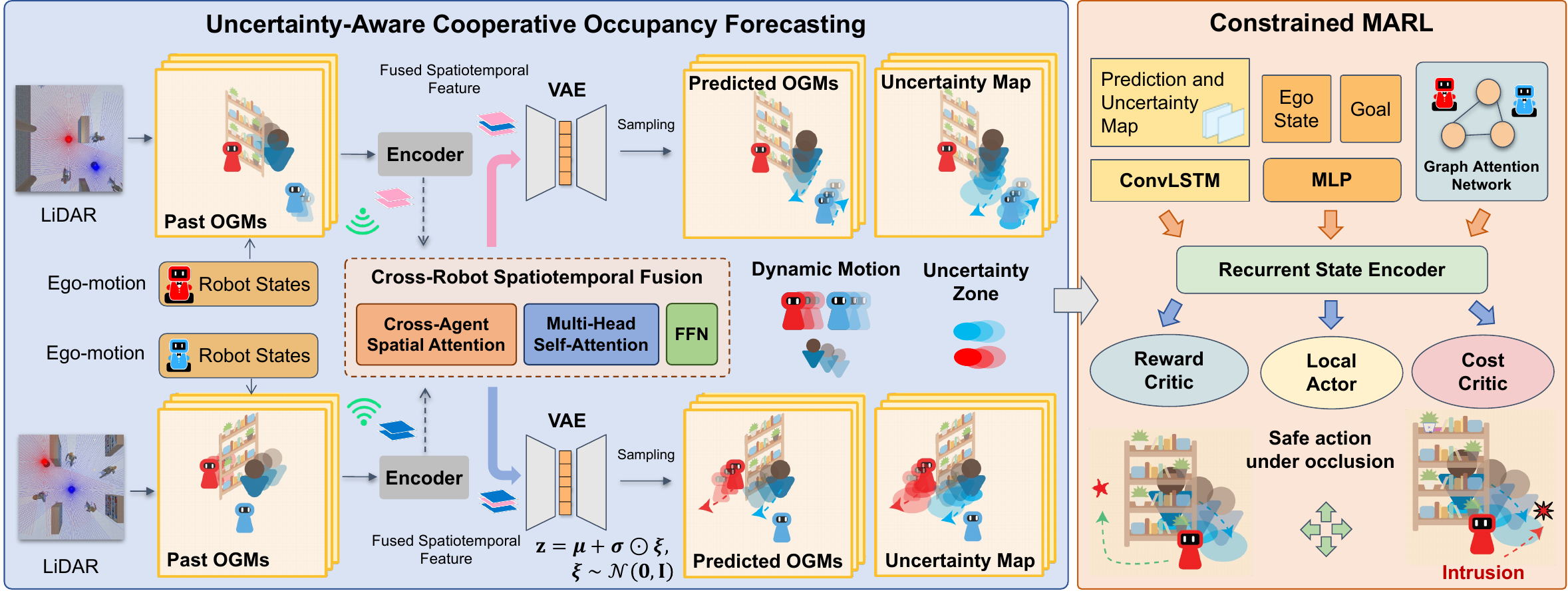}
    \caption{The overall architecture of the CoopUQ for multi-robot social navigation.}
    \vspace{-0.3cm}
    \label{fig:method_overview}
\end{figure}

Figure~\ref{fig:method_overview} provides an overview of the proposed framework. We first encode each robot's LiDAR sequence into spatiotemporal features and employ cross-robot fusion to generate multi-step OGM forecasts and uncertainty estimates, providing a holistic and risk-aware view of the future scene. We then augment the policy observation with the predicted OGMs and uncertainty maps, and guide safe and efficient multi-robot social navigation via constrained MARL.

\subsection{Uncertainty-Aware Cooperative OGM Forecasting}

All robots are equipped with a 2D LiDAR sensor. Let $\mathbf{y}_t^i$ denote the LiDAR measurement of robot $i$ at time $t$, and let $\mathbf{p}_t^i=[p_t^{x_i},p_t^{y_i}]$ denote its position. We convert LiDAR scans into local OGMs using ray-tracing-based occupancy mapping.
LiDAR-derived local OGMs are inherently limited by occlusions and partial observations. A single robot may fail to observe pedestrians behind walls or obstacles, and LiDAR often captures only robot-facing surfaces, making it difficult to infer the complete object extent and shape. These limitations can lead to incomplete future forecasts and situational awareness, and may degrade downstream planning.
To address these challenges, we propose a cooperative OGM predictor that leverages shared information from multiple robots to forecast a more holistic future occupancy representation. Since each robot performs forecasting in its own local frame, the prediction size remains independent of the global environment size. Given the past $\tau$ LiDAR observations $\mathbf{y}_{t-\tau:t}^{1:N}$ and robot positions $\mathbf{p}_{t-\tau:t}^{1:N}$, the cooperative forecasting model predicts the $h$-step OGM sequence $\mathbf{m}_{t+1:t+h}^i\in[0,1]^{h\times m_x\times m_y}$ for each robot $i$. We use $\ell\in\{1,\ldots,h\}$ to index future prediction steps.

\noindent \textbf{Deviation-guided Cross-Robot Spatiotemporal Fusion.}
Each robot converts its scan history into local OGMs and uses its pose history for ego-motion compensation. The compensated OGM history is then encoded by a ConvLSTM encoder into a spatiotemporal feature $\mathbf{F}_t^i$.
For ego robot $i$, features from neighboring robot $k$ are aligned to the ego frame according to the relative pose:
$\hat{\mathbf{F}}_t^{k\rightarrow i}=\Gamma_F(\mathbf{F}_t^k,\mathbf{p}_t^i,\mathbf{p}_t^k)$, where $\Gamma_F(\cdot)$ denotes the learnable spatial transformation.
Instead of applying plain fusion uniformly over all regions, we compute the spatial disagreement between the ego feature and aligned neighboring features, where a large deviation suggests complementary or inconsistent observations caused by occlusion or dynamic agents. This cue is converted into a learnable gate $\mathbf{G}_t^i$ to guide cross-agent spatial attention:
$\mathbf{F}_{\mathrm{cross}}^i =
\mathrm{Attn}_{\mathrm{cross}}(\mathbf{F}_t^i,\{\hat{\mathbf{F}}_t^{k\rightarrow i}\}_{k\neq i};\mathbf{G}_t^i)$.
The gate encourages peer aggregation in uncertain or inconsistent regions while preserving reliable ego features in well-observed areas.
Finally, we apply window-based MHSA to model spatial dependencies in the fused feature and an FFN for channel-wise refinement:
$\mathbf{F}_{\mathrm{fuse}}^i =
\mathrm{FFN}(\mathrm{MHSA}(\mathbf{F}_{\mathrm{cross}}^i))$.

\noindent \textbf{Stochastic Forecasting and Uncertainty Map Estimation.}
Following the stochastic prediction design~\citep{xie2023stochastic, kingma2013auto}, we adopt a variational autoencoder (VAE) to model the future occupancy distribution rather than predicting a deterministic map. 
Given the fused feature $\mathbf{F}_{\mathrm{fuse}}^i$, which encodes the
cross-robot predictive representation of future occupancy from past observations,
the encoder parameterizes a latent distribution as a diagonal Gaussian with mean
$\boldsymbol{\mu}$ and standard deviation $\boldsymbol{\sigma}$, regularized
toward a standard Gaussian prior. Latent variables are sampled via the reparameterization trick and decoded into future occupancy probabilities, capturing the stochasticity of future occupancy induced by dynamic pedestrians and ambiguous motion.
To preserve sharp static geometry while modeling uncertain dynamic regions, we introduce a skip-connected decoder that passes the current OGM directly to the decoding stage. We further sample a separate latent variable for each future step and decode all $h$ future OGMs non-autoregressively, avoiding error accumulation and cost from recursive prediction.
To estimate prediction uncertainty, we draw $M$ samples from the model to obtain $M$ candidate future OGM sequences. Their pixel-wise average forms the predicted OGM sequence $\mathbf{m}_{t+1:t+h}^i$, while the uncertainty map $\mathbf{u}_{t+1:t+h}^i$ is computed using pixel-wise normalized Shannon entropy over the sampled OGMs. Larger uncertainty values often occur near dynamic pedestrians, occluded regions, or ambiguous future motion.

\noindent \textbf{Early and Late Fusion.}
We also consider two alternative fusion strategies for comparison. In the \textit{early fusion} setting, robots share their LiDAR scans with neighboring robots within the communication range. For each ego robot, the scans received from neighboring robots are transformed into the ego frame using relative poses. The ego robot then removes self-hits, converts the aggregated scans into local OGMs, and performs future occupancy forecasting based on the fused sensor input. This strategy preserves low-level geometric information but is more sensitive to communication delay or localization noise.
In the \textit{late fusion} setting, each robot first predicts future OGMs independently using its own observations. The predicted OGMs are then shared with neighboring robots, transformed into the ego frame, and aggregated at the prediction level. This strategy may lose useful intermediate features and can dilute safety-critical predictions.

\subsection{Constrained MARL with Forecasting and Uncertainty}

The observation of robot $i$ is defined as
$\mathbf{o}_t^i = [\mathbf{x}_t^i, \mathbf{m}_{t:t+h}^i, \mathbf{u}_{t:t+h}^i, \mathbf{r}_t^i] \in \mathcal{O}^i$,
where $\mathbf{x}_t^i=[\mathbf{p}_t^i,\mathbf{g}_t^i]$ is the ego state including position and goal.
$\mathbf{m}_{t:t+h}^i$ contains the current OGM and the predicted future OGMs.
$\mathbf{u}_{t:t+h}^i$ contains a zero map at the current step, and the uncertainty maps from future predictions.
Finally, $\mathbf{r}_t^i$ denotes neighboring robot states.
We encode $[\mathbf{m}_{t:t+h}^i,\mathbf{u}_{t:t+h}^i]$ with a ConvLSTM to reason over uncertainty-aware predictions, encode $\mathbf{x}_t^i$ with an MLP, and process $\mathbf{r}_t^i$ with a graph attention network to model robot-robot interactions. The resulting features are concatenated and passed through a GRU for recurrent policy learning.
Beyond using predictions as additional observation channels, CoopUQ leverages the predicted OGM and uncertainty map to explicitly guide risk-aware decision making. We adopt constrained MARL so that the policy learns to maximize task reward while satisfying a safety-cost constraint. The cost is designed to penalize navigation through regions that are predicted to be occupied or highly uncertain.

For robot $i$, we define two local regions around the robot: a safety region $\mathcal{B}_{\text{occ}}^i$ for the predicted OGM and a risk region $\mathcal{B}_{\text{unc}}^i$ for the uncertainty map. A cost is incurred if any cell in the predicted OGM within $\mathcal{B}_{\text{occ}}^i$ exceeds a threshold $\phi_{\text{occ}}$, or if any cell in the uncertainty map within $\mathcal{B}_{\text{unc}}^i$ exceeds an uncertainty threshold $\phi_{\text{unc}}$. The per-robot cost is defined as:
\begin{equation}
\begin{aligned}
    C_i(s_t,\mathbf{a}_t)
    =
    \Delta t \cdot
    \Big[
    &
    \max_{\substack{\ell\in\{0,\ldots,h\} \\ (x,y)\in \mathcal{B}_{\text{occ}}^i}}
    \lambda_{\text{occ}}
    \mathbb{I}
    \left(
    \mathbf{m}_{t+\ell}^{i}(x,y)>\phi_{\text{occ}}
    \right)
    w_d(x,y)
    \\
    +
    &
    \max_{\substack{\ell\in\{0,\ldots,h\} \\ (x,y)\in \mathcal{B}_{\text{unc}}^i}}
    \lambda_{\text{unc}}
    \mathbb{I}
    \left(
    \mathbf{u}_{t+\ell}^{i}(x,y)>\phi_{\text{unc}}
    \right)
    w_d(x,y)
    \Big],
\end{aligned}
\label{eq:cost_function}
\end{equation}
where $\ell=0$ includes the current step, $\mathbb{I}(\cdot)$ is the indicator function, $\lambda_{\text{occ}}$ and $\lambda_{\text{unc}}$ are weighting factors, and $w_d(x,y)$ is a distance-based cell weight with respect to the ego robot. The final cost is scaled by the control duration $\Delta t$. The occupancy term discourages the robot from entering regions likely to be occupied, while the uncertainty term encourages caution in regions where future occupancy is unreliable. This allows the policy to adapt its risk sensitivity under distribution shifts and occlusions, while avoiding unnecessary conservatism in well-observed static regions.

The individual reward for robot $i$ is defined as follows: the robot receives $+10$ if it reaches the goal, i.e., $d_{\text{target}} < D_{\text{goal}}$, $-10$ for collision with static obstacles, $-15$ for collision with other robots, $-20$ for collision with humans, and otherwise receives a dense progress reward $r_{\text{dis}}$. Here, $d_{\text{target}}$ is the Euclidean distance to the goal and $D_{\text{goal}}=\SI{0.5}{m}$ is the goal radius. The dense progress reward is defined as $r_{\text{dis}}=\Delta d \cdot f + r_{\text{time}}$, where $\Delta d=d_{\text{target}}^{t}-d_{\text{target}}^{t+1}$, $f=5$ is the progress scaling factor, and $r_{\text{time}}=-0.1$ is the time penalty. A positive $\Delta d$ indicates progress toward the goal.

We optimize the shared policy using MAPPO-Lagrangian. Following the constrained objective in Eq.~\ref{eq:constrained_opt}, we transform the constrained optimization into the following Lagrangian objective:
\begin{equation}
    \max_{\pi_\theta}\min_{\eta\geq 0}
    L(\pi_\theta,\eta)
    =
    J^R(\pi_\theta)
    -
    \eta
    \left(
    J^C(\pi_\theta)-\epsilon
    \right),
\label{eq:lagrangian_func}
\end{equation}
where $\eta$ is the Lagrange multiplier and $\epsilon$ is the cost limit. We use two separate critics to estimate reward and cost values, respectively. This formulation enables the policy to trade off goal-reaching efficiency and safety under prediction uncertainty.

\section{Experiments}
\label{sec:experiment}

\subsection{Simulation Environment and Dataset}

We build a Unity3D-based multi-robot social navigation simulator with static obstacles, dynamic pedestrians, and multiple robots. Pedestrians are simulated using Unity NavMesh with randomized goals and speeds. For training, three robots navigate in three layouts with 5--10 pedestrians; for in-distribution testing, we use the same layouts with 7--10 pedestrians. OOD evaluation includes unseen layouts, denser crowds with 12--15 pedestrians, and faster pedestrians with speeds up to \SI{2}{m/s}.
We report results with three robots in Table~\ref{tab:iid_nav} and evaluate scalability to five and ten robots in~\refapp{appendix:multi_robot_scalability}.
For OGM forecasting, we collect LiDAR and pose data at \SI{5}{Hz} from the simulator and generate $64\times64$ OGMs with \SI{0.1}{m} resolution. The history window is set to $\tau=5$. The dataset contains 361,152 training and 30,498 testing samples from multi-robot, multi-pedestrian scenarios.

\subsection{Evaluation Metrics and Baselines}

\noindent \textbf{Cooperative Forecasting.}
We evaluate future OGM forecasting over the next 5 steps using metrics from the video forecasting literature~\citep{toyungyernsub2021double}. WMSE measures weighted pixel-wise prediction errors, SSIM reflects the structural consistency of the predicted scene geometry, and IoU measures the overlap between predicted and ground-truth occupied regions. We compare SOGMP++~\citep{xie2023stochastic}, where each robot forecasts OGMs without cooperation, against three cooperation strategies: early fusion, late fusion, and our spatiotemporal (middle) fusion. We further compare performance with clean, \SI{200}{ms} delay, and random localization (Loc.) noise during communication. Details in~\refapp{appendix:evaluation_metrics}.

\noindent \textbf{Social Navigation.}
We evaluate social navigation using metrics, including success rate (SR), collision rate (CR), timeout rate (TR), average navigation time (NT), and path length (PL). We report average acceleration (AA), intrusion time ratio (ITR), and social distance during intrusions (SD)~\citep{liu2023intention}. AA measures trajectory smoothness, while ITR and SD evaluate social awareness around pedestrians. We compare against NavMesh, ORCA~\citep{van2011reciprocal} for reciprocal collision avoidance, SAMARL~\citep{wang2024multi} for learned multi-robot coordination, SCOPE~\citep{xie2025scope} for non-cooperative prediction-aware navigation, and HEIGHT~\citep{liu2026height} for heterogeneous interaction modeling. We augment SAMARL with locally perceived static obstacles and extend HEIGHT to include neighboring robots without peer sharing. For ablation, we compare \textit{Prediction}, which uses predicted OGMs without uncertainty, and \textit{CMARL}, which uses predicted OGMs and uncertainty maps with MAPPO-Lagrangian constraints, under different cooperation modes.

\begin{table*}[!t]
\centering
\scriptsize
\caption{Comparison of OGM forecasting under different cooperation settings.}
\label{tab:ogm_prediction}
\setlength{\tabcolsep}{2.5pt}
\renewcommand{\arraystretch}{0.8}
\resizebox{0.98\textwidth}{!}{\begin{tabular}{l|c|c|ccc|ccc|ccc}
\toprule
\multirow{3}{*}{\textbf{Methods}}
& \multirow{3}{*}{\textbf{Cooperation}}
& \multirow{3}{*}{\textbf{Horizon}}
& \multicolumn{3}{c|}{\textbf{SSIM} $\uparrow$}
& \multicolumn{3}{c|}{\textbf{WMSE} $\downarrow$}
& \multicolumn{3}{c}{\textbf{IoU} $\uparrow$} \\
\cmidrule(lr){4-6} \cmidrule(lr){7-9} \cmidrule(lr){10-12}
& &
& \textbf{Clean}
& \textbf{200ms Delay}
& \textbf{Loc. Noise}
& \textbf{Clean}
& \textbf{200ms Delay}
& \textbf{Loc. Noise}
& \textbf{Clean}
& \textbf{200ms Delay}
& \textbf{Loc. Noise} \\
\midrule

\multirow{2}{*}{SOGMP++ \citep{xie2023stochastic}}
& \multirow{2}{*}{No}
& 1
& 0.75$_{\pm0.00}$
& --
& --
& 0.16$_{\pm0.00}$
& --
& --
& 0.60$_{\pm0.01}$
& --
& -- \\

&
& 5
& 0.71$_{\pm0.00}$
& --
& --
& 0.19$_{\pm0.00}$
& --
& --
& 0.57$_{\pm0.01}$
& --
& -- \\

\midrule

\multirow{6}{*}{\textbf{Ours}}
& \multirow{2}{*}{Early Fusion}
& 1
& 0.86$_{\pm0.00}$
& 0.79$_{\pm0.01}$
& 0.74$_{\pm0.01}$
& 0.07$_{\pm0.00}$
& 0.13$_{\pm0.01}$
& 0.15$_{\pm0.01}$
& 0.73$_{\pm0.01}$
& 0.65$_{\pm0.01}$
& 0.58$_{\pm0.02}$ \\

&
& 5
& 0.80$_{\pm0.00}$
& 0.74$_{\pm0.01}$
& 0.68$_{\pm0.02}$
& 0.12$_{\pm0.00}$
& 0.15$_{\pm0.01}$
& 0.20$_{\pm0.01}$
& 0.68$_{\pm0.01}$
& 0.60$_{\pm0.02}$
& 0.53$_{\pm0.02}$ \\

\cmidrule{2-12}

& \multirow{2}{*}{Late Fusion}
& 1
& 0.81$_{\pm0.01}$
& 0.78$_{\pm0.01}$
& 0.76$_{\pm0.01}$
& 0.12$_{\pm0.00}$
& 0.13$_{\pm0.01}$
& 0.15$_{\pm0.01}$
& 0.63$_{\pm0.01}$
& 0.61$_{\pm0.02}$
& 0.59$_{\pm0.02}$ \\

&
& 5
& 0.76$_{\pm0.01}$
& 0.74$_{\pm0.01}$
& 0.71$_{\pm0.02}$
& 0.14$_{\pm0.01}$
& 0.15$_{\pm0.01}$
& 0.17$_{\pm0.01}$
& 0.60$_{\pm0.01}$
& 0.58$_{\pm0.02}$
& 0.56$_{\pm0.02}$ \\

\cmidrule{2-12}

& \multirow{2}{*}{Middle Fusion}
& 1
& \textbf{0.90$_{\pm0.00}$}
& \textbf{0.88$_{\pm0.01}$}
& \textbf{0.84$_{\pm0.01}$}
& \textbf{0.05$_{\pm0.00}$}
& \textbf{0.06$_{\pm0.01}$}
& \textbf{0.08$_{\pm0.01}$}
& \textbf{0.77$_{\pm0.01}$}
& \textbf{0.74$_{\pm0.01}$}
& \textbf{0.68$_{\pm0.01}$} \\

&
& 5
& \textbf{0.86$_{\pm0.00}$}
& \textbf{0.84$_{\pm0.01}$}
& \textbf{0.80$_{\pm0.01}$}
& \textbf{0.09$_{\pm0.00}$}
& \textbf{0.10$_{\pm0.01}$}
& \textbf{0.12$_{\pm0.01}$}
& \textbf{0.72$_{\pm0.01}$}
& \textbf{0.72$_{\pm0.02}$}
& \textbf{0.64$_{\pm0.02}$} \\

\bottomrule
\end{tabular}}
\vspace{-0.3cm}
\end{table*}

\begin{table*}[!t]
\centering
\footnotesize
\caption{In-distribution navigation performance with three robots.}
\label{tab:iid_nav}
\setlength{\tabcolsep}{2pt}
\renewcommand{\arraystretch}{1.0}
\resizebox{\textwidth}{!}{\begin{tabular}{l|m{1.0cm}<{\centering}|m{1.4cm}<{\centering}|m{1.4cm}<{\centering}m{1.4cm}<{\centering}m{1.4cm}<{\centering}|m{1.4cm}<{\centering}|m{1.5cm}<{\centering}|m{1.5cm}<{\centering}|m{1.5cm}<{\centering}|m{1.5cm}<{\centering}|m{1.5cm}<{\centering}}
\toprule
\multirow{3}{*}{\textbf{Methods}}
& \multirow{3}{*}{\textbf{Coop.}}
& \multirow{3}{*}{\textbf{SR}$\uparrow$}
& \multicolumn{3}{c|}{\textbf{CR}$\downarrow$}
& \multirow{3}{*}{\textbf{TR}$\downarrow$}
& \multirow{3}{*}{\textbf{NT}$\downarrow$}
& \multirow{3}{*}{\textbf{PL}$\downarrow$}
& \multirow{3}{*}{\textbf{AA}$\downarrow$}
& \multirow{3}{*}{\textbf{ITR} (\%)$\downarrow$}
& \multirow{3}{*}{\textbf{SD}$\uparrow$} \\
\cmidrule(lr){4-6}
& & & \textbf{Obstacle} & \textbf{Human} & \textbf{Robots} & & & & & & \\
\midrule

NavMesh & \multirow{7}{*}{No}
& 0.39$_{\pm0.05}$ & \textbf{0.00$_{\pm0.00}$} & 0.41$_{\pm0.06}$ & 0.19$_{\pm0.04}$
& \textbf{0.00$_{\pm0.00}$} & 12.20$_{\pm0.53}$ & 13.89$_{\pm0.62}$ & 0.10$_{\pm0.05}$
& 12.93$_{\pm1.79}$ & 0.29$_{\pm0.03}$ \\

ORCA~\citep{van2011reciprocal} &
& 0.47$_{\pm0.04}$ & 0.17$_{\pm0.02}$ & 0.23$_{\pm0.03}$ & 0.06$_{\pm0.01}$
& 0.06$_{\pm0.02}$ & 10.04$_{\pm0.42}$ & \textbf{9.30$_{\pm0.35}$} & 0.10$_{\pm0.02}$
& 7.63$_{\pm0.48}$ & 0.36$_{\pm0.03}$ \\

SAMARL~\citep{wang2024multi} &
& 0.65$_{\pm0.03}$ & 0.10$_{\pm0.02}$ & 0.12$_{\pm0.02}$ & 0.05$_{\pm0.01}$
& 0.07$_{\pm0.02}$ & 12.98$_{\pm0.46}$ & 14.99$_{\pm0.38}$ & 0.18$_{\pm0.03}$
& 6.60$_{\pm0.42}$ & 0.37$_{\pm0.03}$ \\

SCOPE \cite{xie2025scope} &
& 0.67$_{\pm0.03}$ & 0.10$_{\pm0.02}$ & 0.11$_{\pm0.01}$ & 0.03$_{\pm0.01}$
& 0.09$_{\pm0.02}$ & 13.39$_{\pm0.65}$ & 15.67$_{\pm0.20}$ & 0.18$_{\pm0.02}$
& 5.56$_{\pm0.39}$ & 0.38$_{\pm0.04}$ \\

HEIGHT~\citep{liu2026height} &
& 0.70$_{\pm0.03}$ & 0.07$_{\pm0.01}$ & 0.13$_{\pm0.02}$ & 0.03$_{\pm0.01}$
& 0.06$_{\pm0.01}$ & 12.43$_{\pm0.40}$ & 13.80$_{\pm0.32}$ & 0.14$_{\pm0.02}$
& 4.08$_{\pm0.31}$ & 0.37$_{\pm0.03}$ \\

Prediction &
& 0.71$_{\pm0.01}$ & 0.12$_{\pm0.02}$ & 0.13$_{\pm0.01}$ & 0.02$_{\pm0.00}$
& 0.01$_{\pm0.00}$ & 14.36$_{\pm0.38}$ & 16.94$_{\pm0.26}$ & 0.14$_{\pm0.02}$
& 5.78$_{\pm0.27}$ & 0.34$_{\pm0.03}$ \\

CMARL &
& 0.75$_{\pm0.02}$ & 0.10$_{\pm0.02}$ & 0.10$_{\pm0.03}$ & 0.01$_{\pm0.00}$
& 0.04$_{\pm0.02}$ & 10.82$_{\pm0.18}$ & 13.74$_{\pm0.16}$ & 0.12$_{\pm0.00}$
& 4.17$_{\pm0.26}$ & 0.39$_{\pm0.02}$ \\

\midrule

Prediction & \multirow{2}{*}{Late}
& 0.78$_{\pm0.03}$ & 0.08$_{\pm0.02}$ & 0.10$_{\pm0.02}$ & 0.01$_{\pm0.01}$
& 0.02$_{\pm0.02}$ & 9.89$_{\pm0.18}$ & 12.23$_{\pm0.33}$ & 0.13$_{\pm0.07}$
& 3.96$_{\pm0.19}$ & 0.40$_{\pm0.01}$ \\

CMARL &
& 0.81$_{\pm0.03}$ & 0.07$_{\pm0.02}$ & 0.09$_{\pm0.02}$ & 0.01$_{\pm0.01}$
& 0.01$_{\pm0.02}$ & 10.93$_{\pm0.16}$ & 13.27$_{\pm0.36}$ & 0.12$_{\pm0.05}$
& 3.37$_{\pm0.16}$ & 0.42$_{\pm0.02}$ \\

\midrule

Prediction & \multirow{2}{*}{Early}
& 0.77$_{\pm0.02}$ & 0.11$_{\pm0.01}$ & 0.06$_{\pm0.02}$ & 0.01$_{\pm0.01}$
& 0.05$_{\pm0.02}$ & 9.94$_{\pm0.19}$ & 12.55$_{\pm0.33}$ & 0.14$_{\pm0.06}$
& 2.68$_{\pm0.36}$ & 0.43$_{\pm0.01}$ \\

CMARL &
& 0.82$_{\pm0.02}$ & 0.07$_{\pm0.01}$ & 0.05$_{\pm0.01}$ & \textbf{0.00$_{\pm0.01}$}
& 0.06$_{\pm0.02}$ & 11.55$_{\pm0.36}$ & 14.30$_{\pm0.47}$ & 0.12$_{\pm0.05}$
& 2.37$_{\pm0.19}$ & 0.43$_{\pm0.01}$ \\

\midrule

Prediction & \multirow{2}{*}{Middle}
& 0.84$_{\pm0.02}$ & 0.06$_{\pm 0.01}$ & 0.05$_{\pm0.01}$ & 0.01$_{\pm0.01}$
& 0.04$_{\pm0.02}$ & 10.46$_{\pm0.09}$ & 12.08$_{\pm0.29}$ & 0.11$_{\pm0.03}$
& 2.17$_{\pm0.26}$ & 0.42$_{\pm0.01}$ \\

\textbf{CMARL} &
& \textbf{0.90$_{\pm0.03}$} & 0.03$_{\pm0.01}$ & \textbf{0.03$_{\pm0.01}$} & \textbf{0.00$_{\pm0.00}$}
& 0.03$_{\pm0.01}$ & \textbf{9.80$_{\pm0.15}$} & 11.98$_{\pm0.29}$ & 0.13$_{\pm0.05}$
& \textbf{2.07$_{\pm0.19}$} & \textbf{0.45$_{\pm0.03}$} \\

\bottomrule
\end{tabular}}
\vspace{-0.3cm}
\end{table*}

\begin{figure*}[t]
    \centering
    \includegraphics[width=\linewidth]{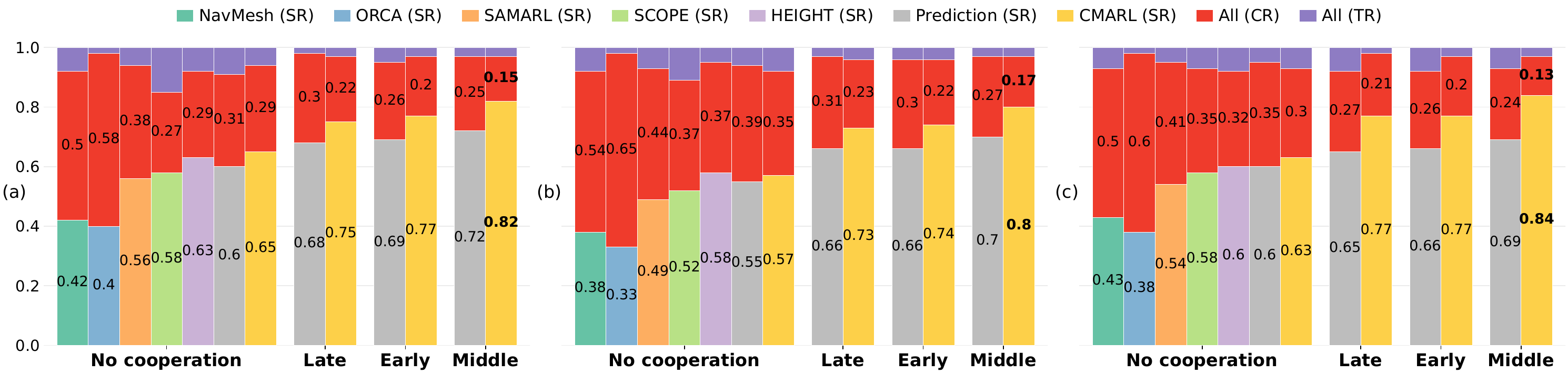}
    \vspace{-0.3cm}
    \caption{OOD test results in (a) unseen layouts, (b) denser crowds, and (c) rushing pedestrians.}
    \label{fig:ood_navigation}
\end{figure*}

\subsection{Quantitative Results}

\noindent \textbf{Cooperative Forecasting.}
Table~\ref{tab:ogm_prediction} reports the mean and standard deviation over three test seeds. Under clean inputs, SOGMP++ performs worst across horizons, showing the limitation of individual-only forecasting under partial observability. All fusion strategies improve prediction quality by incorporating complementary views, with early fusion outperforming late fusion by reasoning over a more complete context before prediction.
CoopUQ (middle fusion) achieves the best across all metrics by preserving richer spatiotemporal context. At the 5-step horizon, it improves SSIM from 0.71 to 0.86 and IoU from 0.57 to 0.72 over SOGMP++, yielding relative gains of 21.1\% and 26.3\%.
Under a \SI{200}{ms} delay at the same 5-step horizon, it shows mild degradation, with a 2.3\% SSIM drop and unchanged IoU, compared with larger drops for early and late fusion.
Under localization noise, it retains the highest SSIM, with a 7.0\% drop versus 15.0\% and 6.6\% for early and late fusion, as the learnable spatial transformation mitigates small misalignments. Overall, CoopUQ improves forecasting performance and remains robust.

\noindent \textbf{In-distribution Navigation.}
Table~\ref{tab:iid_nav} summarizes in-distribution navigation performance with three robots. NavMesh avoids static obstacles with privileged map access, but suffers the lowest SR and high human/robot collision rates in dense dynamic crowds. SCOPE~\citep{xie2025scope} improves over NavMesh but is limited by single-robot prediction.
ORCA suffers from frequent human and obstacle collisions, while SAMARL and HEIGHT improve SR but remain limited by local observations. ORCA attains the shortest paths, whereas CoopUQ achieves the highest SR, lowest human CR and ITR, shortest NT, and largest SD. This contrast highlights the importance of considering navigation efficiency together with success and safety.
Cooperative forecasting consistently improves navigation performance. Middle fusion achieves the strongest prediction-only result, reaching 0.84 SR and reducing human collision rate to 0.05, indicating that spatiotemporal fusion better preserves cross-robot context. Adding uncertainty-aware CMARL further improves safety. Without cooperation, SR increases from 0.71 to 0.75, and the human collision rate drops from 0.13 to 0.10. Lower ITR and higher SD further suggest that uncertainty maps help the policy avoid ambiguous high-risk regions while maintaining efficient motion.
Overall, CoopUQ improves SR from 0.75 to 0.90 and reduces the human collision rate from 0.10 to 0.03 compared with non-cooperative CMARL. These results show that uncertainty-aware cooperative forecasting and risk-aware control jointly improve safety, efficiency, and social compliance in constrained dynamic environments.

\noindent \textbf{OOD Navigation.}
Fig.~\ref{fig:ood_navigation} shows the OOD navigation results. All learning-based methods degrade under distribution shifts, while CoopUQ achieves the highest SR across unseen layouts, denser crowds, and faster pedestrian motion. In unseen layouts, CoopUQ maintains the highest SR, indicating improved transferability from cooperative forecasting. Under increased pedestrian density, SCOPE drops by 22.4\% in SR, whereas CoopUQ drops by only 11.1\%. In the rushing-pedestrian setting, CoopUQ shows an SR drop of 6.7\%, suggesting better robustness to sudden human motion. The widening performance gap shows that CoopUQ improves reliability in unfamiliar environments or unpredictable human behaviors, where forecasting becomes challenging.

\subsection{Qualitative Results}
\begin{figure*}[!t]
    \centering
    \includegraphics[width=\linewidth]{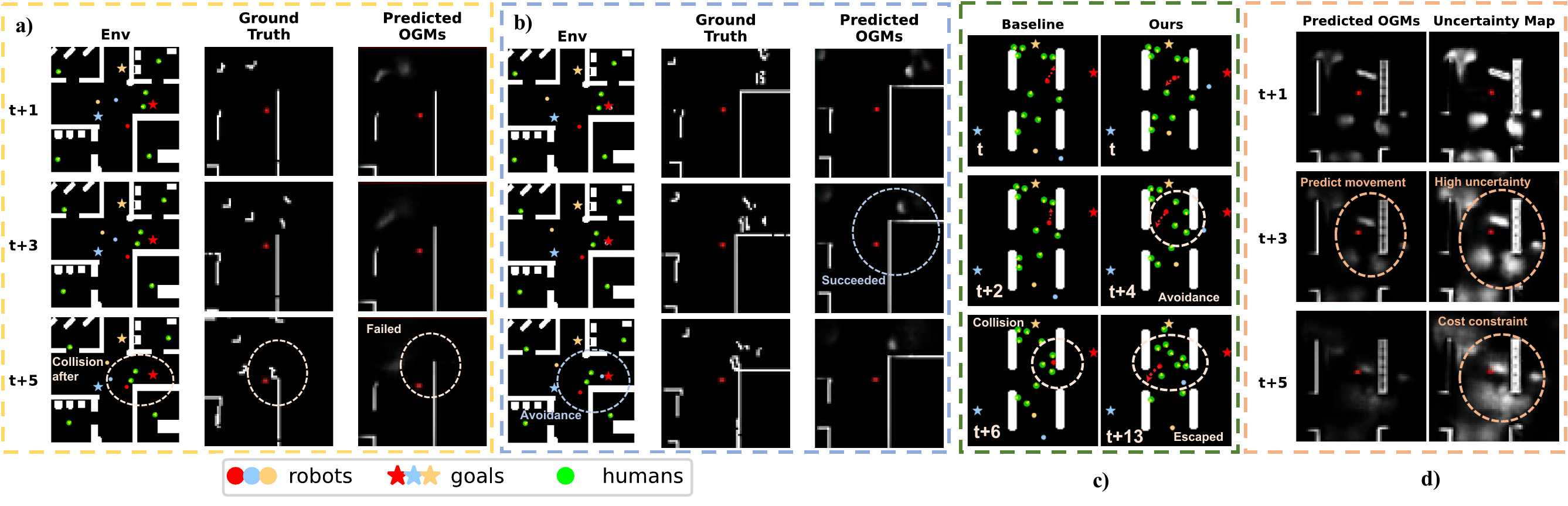}
    \vspace{-0.3cm}
    \caption{(a) Prediction w/o cooperation: The red robot reacts only after the human fully emerges into its view, resulting in a sharp brake and collision. (b) Prediction w/ cooperation: The red robot decelerates and changes direction earlier to avoid a collision. (c) Key frames of navigation w/o and w/ cooperation. (d) Prediction and uncertainty map from ours at step $t$. The red ego robot proactively avoids entering high-uncertainty areas due to cost constraints, thus preventing potential collisions.}
    \label{fig:qualitative_navigation}
    \vspace{-0.3cm}
\end{figure*}

\noindent \textbf{Cooperative Forecasting.}
Figure~\ref{fig:qualitative_navigation} (a) and (b) compare the predicted OGMs of SCOPE and CoopUQ in a scenario where three robots navigate through human crowds and diverse obstacle geometries. As the red ego robot approaches a wall corner with limited visibility, a pedestrian emerges from behind the corner and crosses its path. SCOPE reacts only after the pedestrian becomes fully visible, leading to a collision. In contrast, CoopUQ leverages shared spatiotemporal features from neighboring robots to recover occluded regions and predict more complete scene geometry and dynamic occupancy, enabling the ego robot to detect the risk earlier and proactively avoid the collision.

\noindent \textbf{Social Navigation.}
Figure~\ref{fig:qualitative_navigation}(c) and (d) compare SCOPE and CoopUQ in a constrained scene with dense, randomly moving pedestrians. SCOPE reacts late to pedestrians and obstacles, brakes abruptly, and can become trapped due to limited field of view and inaccurate obstacle estimation.
In contrast, CoopUQ leverages shared spatiotemporal features to better forecast obstacle geometry and pedestrian motion, enabling earlier deceleration and proactive direction changes. Its uncertainty map highlights ambiguous regions around pedestrians, allowing the constrained policy to avoid high-risk areas before collisions become imminent. Overall, uncertainty-aware cooperative forecasting and constrained control lead to safer, more efficient, and socially compliant navigation.

\subsection{Real Robot Experiments}
We deploy our method on two ROSMASTER X3 robots. Each uses an Ultra-Wideband (UWB) for localization, with measurements fused via a Kalman filter. Robots connect to a laptop with an NVIDIA RTX 3070 GPU for inference. The policy is trained in Unity simulation with holonomic dynamics and deployed to the robot without fine-tuning.
We test in four everyday environments with one to three humans walking around. CoopUQ achieves an average SR of 0.94 across 48 trials. 
SCOPE obtains an average SR of 0.58, as it reacts only after pedestrians become fully visible in occluded scenarios, leading to collisions. Detailed quantitative results are provided in~\refapp{appendix:real_robot_experiments}.

\section{Limitations and Future Work}
This work focuses on homogeneous 2D LiDAR-equipped robots and short-horizon occupancy forecasting. This may limit CoopUQ under heterogeneous sensing or scenarios requiring longer-term human intent reasoning. Future work will explore heterogeneous robot teams with richer sensor modalities. The uncertainty map also does not explicitly capture semantic information. Incorporating semantic occupancy from LiDAR is a promising direction. Finally, while we evaluate OOD performance under unseen layouts, denser crowds, and faster pedestrian motion, real-world distribution shifts are broader. Future work will test more diverse environments and pedestrian behaviors.

\section{Conclusion}
\label{sec:conclusion}

We introduce CoopUQ, a multi-robot social navigation framework that bridges uncertainty-aware cooperative forecasting with robust decision making. CoopUQ uses cross-robot spatiotemporal fusion to produce accurate occupancy forecasts and incorporates prediction uncertainty into a constrained MARL policy. This enables robots to avoid ambiguous high-risk regions while maintaining efficiency in well-observed areas. Extensive Unity3D simulations and real-world experiments demonstrate improved safety, efficiency, and robustness across in-distribution and OOD scenarios.

\bibliography{example}
\clearpage

\appendix
\textbf{\Large{Appendix}}
\vspace{5pt}

\section{Additional Experimental Results}

\subsection{Real Robot Experiments}
\label{appendix:real_robot_experiments}

\begin{figure*}[h]
    \centering
    \includegraphics[width=\linewidth]{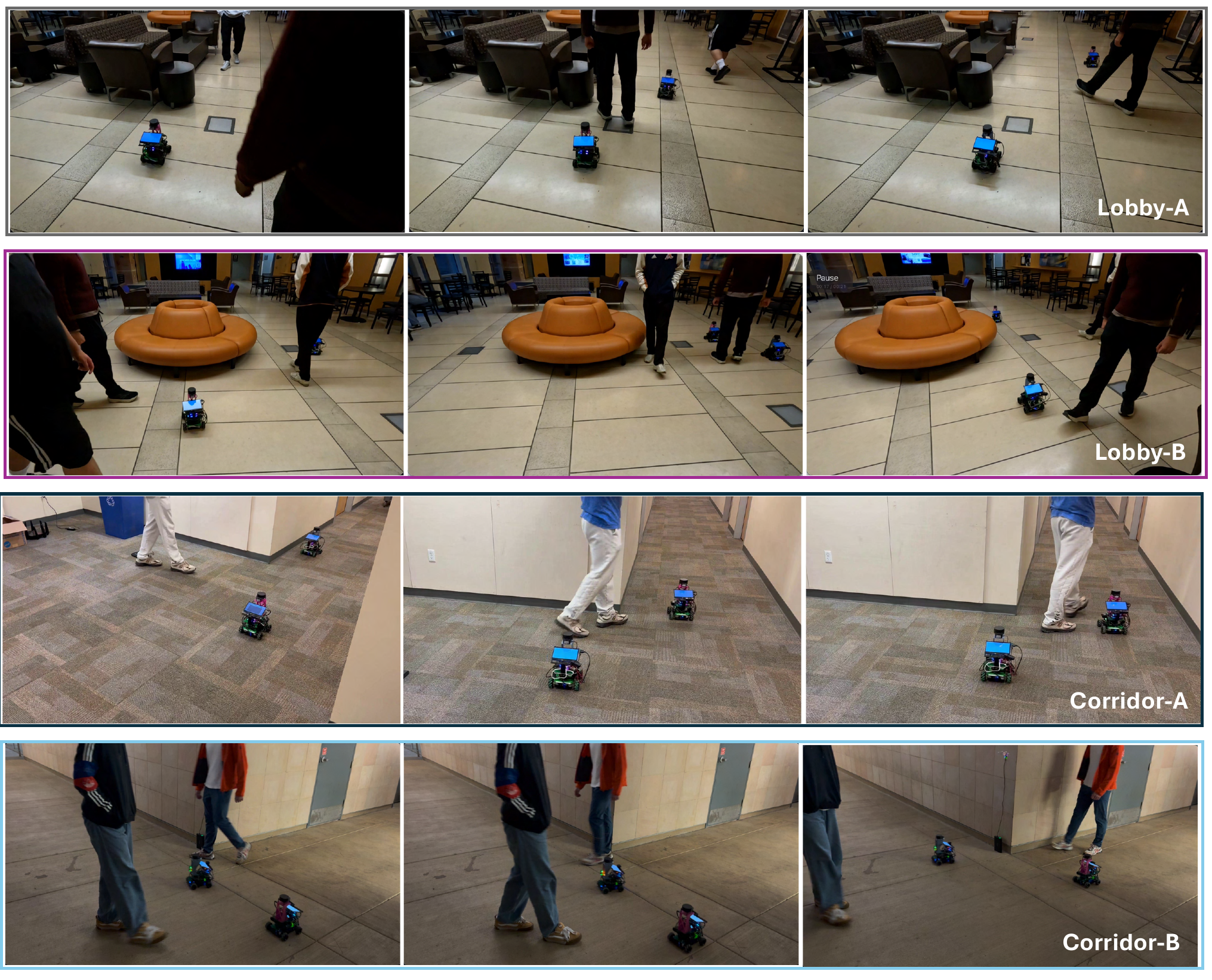}
    \caption{We evaluate CoopUQ in four indoor environments with one to three pedestrians walking in the scene. In total, 48 trials are conducted (4 per environment and pedestrian count). Across all runs, CoopUQ achieves an average success rate of 0.94. The success rate with one, two, and three pedestrians is 1.0, 0.94, and 0.88, respectively.}
    \label{fig:real_world_environments}
\end{figure*}

\begin{figure*}[h]
    \centering
    \includegraphics[width=\linewidth]{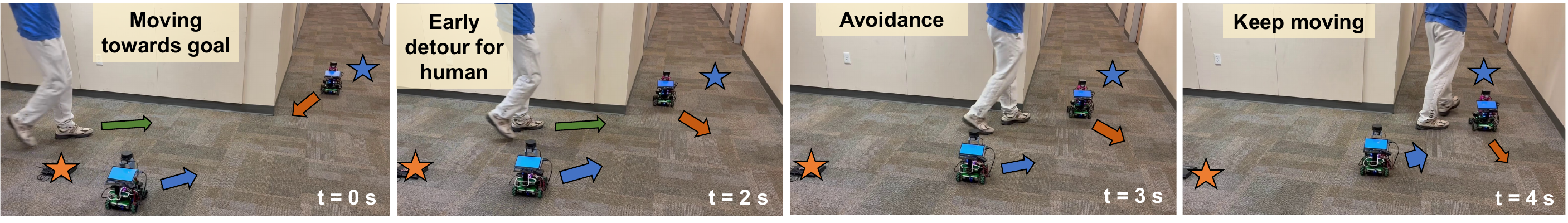}
    \caption{The upper orange robot navigates toward its goal while detouring early to avoid colliding with a pedestrian emerging from occlusion. The demo is available in supplementary videos.}
    \label{fig:real_world_early_detour}
\end{figure*}

We develop a modularized ROS~2 system consisting of prediction and decision making nodes. 
The robot is equipped with a 2D LiDAR (RPLIDAR-A1) operating at a scanning frequency of approximately \SI{6}{Hz}. 
The prediction node converts raw LiDAR measurements into occupancy grid maps and performs online occupancy forecasting using the same model architecture, parameters, and weights as those employed during training in simulator without any additional training in the real world.
The decision making node receives the predicted OGMs from the prediction node and computes action commands in the form of planar velocity $(v_x, v_y)$. These commands are executed using the onboard controller integrated into the ROSMASTER X3 platform by publishing to the \texttt{/cmd\_vel} topic. 

\begin{table*}[h]
    \centering
    \caption{Real-world navigation performance across different environments and pedestrian densities. Results are averaged over 4 trials for each environment and pedestrian setting.}
    \label{tab:real_world_results}
    \resizebox{1.0\linewidth}{!}{
    \setlength{\tabcolsep}{2pt}
    \renewcommand{\arraystretch}{1.3}
    \begin{tabular}{l|c|ccc|ccc|ccc}
        \toprule
        \multirow{3}{*}{\textbf{Method}} 
        & \multirow{3}{*}{\textbf{Environment}} 
        & \multicolumn{9}{c}{\textbf{Pedestrians Count}} \\
        \cline{3-11}
        & & \multicolumn{3}{c|}{\textbf{1}} 
          & \multicolumn{3}{c|}{\textbf{2}} 
          & \multicolumn{3}{c}{\textbf{3}} \\
        \cline{3-11}
        & & \textbf{Success} $\uparrow$ & \textbf{Collision} $\downarrow$ & \textbf{Timeout}
          & \textbf{Success} $\uparrow$ & \textbf{Collision} $\downarrow$ & \textbf{Timeout}
          & \textbf{Success} $\uparrow$ & \textbf{Collision} $\downarrow$ & \textbf{Timeout} \\
        \midrule

        \multirow{4}{*}{SCOPE~\citep{xie2025scope}}
        & Lobby-A     & 1.00 & 0.00 & 0.00 & 0.75 & 0.25 & 0.00 & 0.50 & 0.25 & 0.25 \\
        & Lobby-B     & 0.75 & 0.00 & 0.25 & 0.50 & 0.25 & 0.25 & 0.25 & 0.50 & 0.25 \\
        & Corridor-A  & 0.75 & 0.25 & 0.00 & 0.50 & 0.50 & 0.00 & 0.25 & 0.50 & 0.25 \\
        & Corridor-B  & 0.75 & 0.25 & 0.00 & 0.50 & 0.50 & 0.00 & 0.50 & 0.50 & 0.00 \\

        \midrule

        \multirow{4}{*}{\textbf{CoopUQ}}
        & Lobby-A     & 1.00 & 0.00 & 0.00 & 1.00 & 0.00 & 0.00 & 1.00 & 0.00 & 0.00 \\
        & Lobby-B     & 1.00 & 0.00 & 0.00 & 1.00 & 0.00 & 0.00 & 0.75 & 0.00 & 0.25 \\
        & Corridor-A  & 1.00 & 0.00 & 0.00 & 0.75 & 0.00 & 0.25 & 0.75 & 0.25 & 0.00 \\
        & Corridor-B  & 1.00 & 0.00 & 0.00 & 1.00 & 0.00 & 0.00 & 1.00 & 0.00 & 0.00 \\

        \midrule
        \multirow{2}{*}{Average}
        & SCOPE & 0.81 & 0.13 & 0.06 & 0.56 & 0.38 & \textbf{0.06} & 0.38 & 0.44 & 0.19 \\
        & \textbf{CoopUQ}     & \textbf{1.00} & \textbf{0.00} & \textbf{0.00}
                     & \textbf{0.94} & \textbf{0.00} & \textbf{0.06}
                     & \textbf{0.88} & \textbf{0.06} & \textbf{0.06} \\
        \bottomrule
    \end{tabular}
    }
\end{table*}

As shown in Figure~\ref{fig:real_world_environments} and Figure~\ref{fig:real_world_early_detour}, we further enhance the real-world experiments to four distinct everyday environments with pedestrian density variations. The corridor environment has more limited free space, and the lobby environment has obstacles with more diverse shapes. For each environment, experiments are conducted with one, two, and three pedestrians walking in the scene, and each configuration is repeated four times, resulting in a total of 48 trials. During each experiment, each of the two ROSMASTER X3 robots navigates toward a predetermined target. In most cases, each robot’s goal is set to the initial position of the other robot.
Pedestrians were not instructed to avoid the robots.
Detailed real-world quantitative results are summarized in Table~\ref{tab:real_world_results}. Across all runs, CoopUQ achieves an average success rate of 0.94, exhibiting smooth trajectories and early detouring behaviors under occlusion. The success rate with one, two, and three pedestrians is 1.0, 0.94, and 0.88, respectively. SCOPE obtains an average success rate of 0.58, as it reacts only after pedestrians become fully visible in occluded scenarios, often leading to collisions.
We observe that collisions mainly occur in narrow corridor scenarios with higher pedestrian density, where the available free space is highly constrained. When pedestrians surround the robot and block the feasible space, the robot may collide with nearby walls. Timeout cases are primarily attributed to conservative behaviors, in which the robot slows down or goes back and forth to maintain safety when the surroundings have cluttered obstacles (e.g., tables, sofas, and chairs) and pedestrians are continuously walking near the robot.
These observations suggest several directions for future work. First, incorporating a stop-and-wait strategy could help reduce collisions in highly constrained environments. Second, adopting risk modulation under prolonged stalling, or integrating longer-horizon pedestrian intent prediction, may further improve navigation efficiency under denser spaces.

\subsection{Scalability across Robot Team Sizes}
\label{appendix:multi_robot_scalability}

Table~\ref{tab:scaling_baselines} extends the three-robot evaluation in Table~\ref{tab:iid_nav} to five and ten robots without retraining. CoopUQ maintains a local OGM for each robot and aggregates features from nearby teammates, allowing cooperative forecasting to accommodate changing local neighborhoods as the team size grows. It retains the highest SR and the lowest human CR, robot CR, and ITR at both team sizes, demonstrating effective transfer to larger teams. The increased navigation times and path lengths reflect additional local avoidance maneuvers needed as more robots share the same space. Meanwhile, lower SD indicates tighter pedestrian clearance in these more crowded settings.

\begin{table}[htbp]
\centering
\scriptsize
\caption{Navigation performance with three, five and ten robots. Values are reported as ID/OOD. Three-robot OOD results are averaged equally over the three scenarios. All learned policies are trained with three robots and evaluated zero-shot with five and ten robots.}
\label{tab:scaling_baselines}
\setlength{\tabcolsep}{3pt}
\resizebox{\textwidth}{!}{\begin{tabular}{clcccccccccc}
\toprule
\multirow[c]{2}{*}[-0.8ex]{\textbf{Robots}} & \multirow[c]{2}{*}[-0.8ex]{\textbf{Method}} & \multirow[c]{2}{*}[-0.8ex]{\textbf{SR}$\uparrow$} & \multicolumn{3}{c}{\textbf{CR}$\downarrow$} & \multirow[c]{2}{*}[-0.8ex]{\textbf{TR}$\downarrow$} & \multirow[c]{2}{*}[-0.8ex]{\textbf{NT}$\downarrow$} & \multirow[c]{2}{*}[-0.8ex]{\textbf{PL}$\downarrow$} & \multirow[c]{2}{*}[-0.8ex]{\textbf{AA}$\downarrow$} & \multirow[c]{2}{*}[-0.8ex]{\textbf{ITR} (\%)$\downarrow$} & \multirow[c]{2}{*}[-0.8ex]{\textbf{SD}$\uparrow$} \\
\cmidrule(lr){4-6}
& & & Obstacle & Human & Robots & & & & & & \\
\midrule
\multirow{5}{*}{Three}
& ORCA~\citep{van2011reciprocal} & 0.47/0.37 & 0.17/0.15 & 0.23/0.38 & 0.06/0.08 & 0.06/\textbf{0.02} & 10.04/\textbf{10.54} & \textbf{9.30}/\textbf{11.12} & 0.10/0.11 & 7.63/11.80 & 0.36/0.34 \\
& SAMARL~\citep{wang2024multi} & 0.65/0.53 & 0.10/0.14 & 0.12/0.21 & 0.05/0.06 & 0.07/0.06 & 12.98/13.45 & 14.99/15.57 & 0.18/0.20 & 6.60/10.43 & 0.37/0.34 \\
& SCOPE~\citep{xie2025scope} & 0.67/0.56 & 0.10/0.10 & 0.11/0.18 & 0.03/0.05 & 0.09/0.11 & 13.39/14.79 & 15.67/16.92 & 0.18/0.21 & 5.56/7.56 & 0.38/0.36 \\
& HEIGHT~\citep{liu2026height} & 0.70/0.60 & 0.07/0.08 & 0.13/0.19 & 0.03/0.05 & 0.06/0.07 & 12.43/13.49 & 13.80/14.11 & 0.14/0.15 & 4.08/8.02 & 0.37/0.36 \\
& CoopUQ & \textbf{0.90}/\textbf{0.82} & \textbf{0.03}/\textbf{0.04} & \textbf{0.03}/\textbf{0.10} & \textbf{0.00}/\textbf{0.01} & \textbf{0.03}/0.03 & \textbf{9.80}/10.88 & 11.98/13.82 & 0.13/0.17 & \textbf{2.07}/\textbf{4.47} & \textbf{0.45}/\textbf{0.43} \\
\midrule
\multirow{5}{*}{Five}
& ORCA~\citep{van2011reciprocal} & 0.44/0.34 & 0.18/0.13 & 0.20/0.37 & 0.12/0.14 & 0.05/\textbf{0.02} & \textbf{9.97}/\textbf{10.73} & \textbf{9.36}/\textbf{11.26} & 0.10/0.10 & 7.04/11.14 & 0.37/0.34 \\
& SAMARL~\citep{wang2024multi} & 0.62/0.51 & 0.11/0.14 & 0.10/0.18 & 0.08/0.11 & 0.08/0.06 & 13.07/13.62 & 15.21/15.89 & 0.18/0.20 & 6.03/9.26 & 0.36/0.34 \\
& SCOPE~\citep{xie2025scope} & 0.66/0.56 & 0.10/0.11 & 0.12/0.18 & 0.06/0.06 & 0.07/0.09 & 13.35/14.80 & 15.88/16.89 & 0.18/0.21 & 6.31/7.01 & 0.38/0.36 \\
& HEIGHT~\citep{liu2026height} & 0.70/0.61 & 0.08/0.09 & 0.13/0.18 & 0.04/0.06 & 0.04/0.05 & 12.78/13.40 & 14.38/14.05 & 0.15/0.15 & 3.90/7.09 & 0.37/0.36 \\
& CoopUQ & \textbf{0.88}/\textbf{0.81} & \textbf{0.03}/\textbf{0.05} & \textbf{0.05}/\textbf{0.10} & \textbf{0.01}/\textbf{0.01} & \textbf{0.02}/\textbf{0.02} & 10.62/11.71 & 12.62/14.86 & 0.13/0.17 & \textbf{2.26}/\textbf{4.42} & \textbf{0.44}/\textbf{0.42} \\
\midrule
\multirow{5}{*}{Ten}
& ORCA~\citep{van2011reciprocal} & 0.38/0.30 & 0.14/0.12 & 0.20/0.34 & 0.24/0.24 & 0.03/\textbf{0.01} & \textbf{9.99}/\textbf{10.26} & \textbf{9.21}/\textbf{10.86} & 0.10/0.11 & 7.03/10.72 & 0.34/0.33 \\
& SAMARL~\citep{wang2024multi} & 0.53/0.44 & 0.11/0.14 & 0.07/0.17 & 0.21/0.20 & 0.07/0.06 & 13.85/13.77 & 16.00/16.03 & 0.19/0.21 & 5.84/9.47 & 0.37/0.34 \\
& SCOPE~\citep{xie2025scope} & 0.62/0.52 & 0.12/0.11 & 0.12/0.20 & 0.10/0.11 & 0.04/0.06 & 13.75/14.22 & 16.49/16.38 & 0.18/0.21 & 7.60/7.92 & 0.37/0.36 \\
& HEIGHT~\citep{liu2026height} & 0.57/0.56 & 0.14/0.14 & 0.19/0.19 & 0.08/0.08 & \textbf{0.01}/0.03 & 12.78/13.57 & 14.61/14.37 & 0.15/0.16 & 5.48/7.08 & 0.36/0.36 \\
& CoopUQ & \textbf{0.84}/\textbf{0.78} & \textbf{0.04}/\textbf{0.06} & \textbf{0.05}/\textbf{0.10} & \textbf{0.02}/\textbf{0.02} & 0.04/0.03 & 12.85/13.55 & 16.16/17.16 & 0.16/0.18 & \textbf{2.95}/\textbf{4.62} & \textbf{0.42}/\textbf{0.40} \\
\bottomrule
\end{tabular}}
\end{table}

\subsection{Effectiveness of Deviation-guided Cross-Robot Spatiotemporal Fusion}
\label{appendix:middle_fusion_ablation}

\begin{table}[H]
\centering
\scriptsize
\caption{Ablation study on deviation-guided cross-robot spatiotemporal fusion for OGM forecasting.}
\label{tab:middle_fusion_ablation}
\setlength{\tabcolsep}{4pt}
\renewcommand{\arraystretch}{0.9}
\resizebox{1.0\textwidth}{!}{
\begin{tabular}{lcccccc}
\toprule
& \multicolumn{3}{c}{\textbf{1-step}} & \multicolumn{3}{c}{\textbf{5-step}} \\
\cmidrule(lr){2-4} \cmidrule(lr){5-7}
\textbf{Method} & SSIM $\uparrow$ & WMSE $\downarrow$ & IoU $\uparrow$
                & SSIM $\uparrow$ & WMSE $\downarrow$ & IoU $\uparrow$ \\
\midrule
No Cooperation                   & $0.75 \pm 0.00$ & $0.16 \pm 0.00$ & $0.60 \pm 0.01$
                                 & $0.71 \pm 0.00$ & $0.19 \pm 0.00$ & $0.57 \pm 0.01$ \\
Early Fusion                     & $0.86 \pm 0.00$ & $0.07 \pm 0.00$ & $0.73 \pm 0.01$
                                 & $0.80 \pm 0.00$ & $0.12 \pm 0.00$ & $0.68 \pm 0.01$ \\
Late Fusion                      & $0.81 \pm 0.01$ & $0.12 \pm 0.00$ & $0.63 \pm 0.01$
                                 & $0.76 \pm 0.01$ & $0.14 \pm 0.01$ & $0.60 \pm 0.01$ \\
Middle Fusion w/o Deviation-guided & $0.87 \pm 0.01 $ & $0.08 \pm 0.01 $ & $0.74 \pm 0.01$
                                 & $0.83 \pm 0.01$ & $0.11 \pm 0.01$ & $0.71 \pm 0.01$ \\
Middle Fusion w/ Deviation-guided & $\mathbf{0.90 \pm 0.00}$ & $\mathbf{0.05 \pm 0.00}$ & $\mathbf{0.77 \pm 0.01}$
                                 & $\mathbf{0.86 \pm 0.00}$ & $\mathbf{0.09 \pm 0.00}$ & $\mathbf{0.72 \pm 0.01}$ \\
\bottomrule
\end{tabular}
}
\end{table}

Table~\ref{tab:middle_fusion_ablation} isolates the effect of the deviation gate within middle fusion. Compared with plain middle fusion, which fuses aligned neighboring features uniformly, the deviation-guided fusion consistently improves all metrics at both horizons. Absolute improvements in WMSE and IoU are larger at the 1-step horizon, while SSIM improves by $0.03$ at both horizons. At the 5-step horizon, SSIM increases from $0.83$ to $0.86$, WMSE decreases from $0.11$ to $0.09$, and IoU increases from $0.71$ to $0.72$. These results show that spatial deviation is an effective cue for identifying regions where cross-robot information is useful. By strengthening aggregation in occluded or inconsistent areas while preserving reliable ego features, the deviation gate produces more accurate forecasts.

\subsection{Effectiveness of CMARL}
\label{appendix:cmarl_abaltion}

\begin{wrapfigure}{r}{0.45\linewidth}
\vspace{-\intextsep}
\centering
\setlength{\abovecaptionskip}{0pt}
\includegraphics[width=\linewidth]{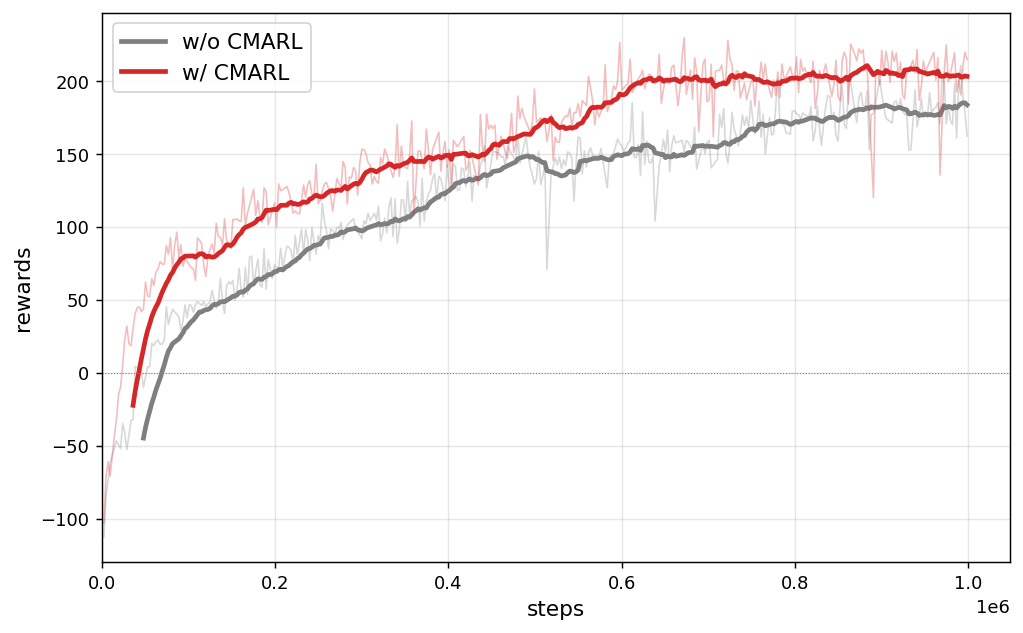}
\vspace{-0.4cm}
\caption{Training reward comparison with and without CMARL. The CMARL policy
achieves faster convergence and higher final rewards, indicating that cost constraints do not merely improve safety but also facilitate efficient policy learning.}
\label{fig:cmarl_training_rewards}
\vspace{-0.3cm}
\end{wrapfigure}

Table~\ref{tab:cmarl_ablation} compares three ways of using uncertainty across cooperation strategies. Prediction uses only predicted OGMs, while Prediction \& Uncertainty further adds uncertainty maps as observations without explicit constraints. This yields only marginal gains, suggesting that uncertainty is difficult to exploit from task rewards alone. In contrast, CMARL further constrains safety cost based on predicted occupancy and uncertainty for MAPPO-Lagrangian optimization, leading to the best performance. The lower ITR and higher SD further show that CMARL guides robots away from ambiguous regions while avoiding predicted occupied areas.
Figure~\ref{fig:cmarl_training_rewards} shows that CMARL converges faster and reaches higher final rewards, indicating improved training efficiency.

\begin{table}[h]
\centering
\scriptsize
\caption{Ablation study on CMARL under in-distribution navigation.}
\label{tab:cmarl_ablation}
\setlength{\tabcolsep}{3pt}
\renewcommand{\arraystretch}{0.9}
\resizebox{1.0\textwidth}{!}{\begin{tabular}{llccccccc}
\toprule
\textbf{Cooperation} & \textbf{Policy} & \textbf{SR} $\uparrow$ & \textbf{Obstacle CR} $\downarrow$ & \textbf{Human CR} $\downarrow$ & \textbf{Robots CR} $\downarrow$ & \textbf{TR} $\downarrow$ & \textbf{ITR} (\%) $\downarrow$ & \textbf{SD} $\uparrow$ \\
\midrule
\multirow{3}{*}{No Cooperation}
 & Prediction              & $0.71 \pm 0.01$ & $0.12 \pm 0.02$ & $0.13 \pm 0.01$ & $0.02 \pm 0.00$ & $\mathbf{0.01 \pm 0.00}$ & $5.78 \pm 0.27$ & $0.34 \pm 0.03$ \\
 & Prediction \& Uncertainty & $0.72 \pm 0.01$ & $0.11 \pm 0.02$ & $0.12 \pm 0.01$ & $0.02 \pm 0.01 $ & $0.03 \pm 0.00$ & $5.26 \pm 0.37$ & $0.35 \pm 0.01$ \\
 & CMARL                   & $0.75 \pm 0.02$ & $0.10 \pm 0.02$ & $0.10 \pm 0.03$ & $0.01 \pm 0.00$ & $0.04 \pm 0.02$ & $4.17 \pm 0.26$ & $0.39 \pm 0.02$ \\
\midrule
\multirow{3}{*}{Late Fusion}
 & Prediction              & $0.78 \pm 0.03$ & $0.08 \pm 0.02$ & $0.10 \pm 0.02$ & $0.01 \pm 0.01$ & $0.02 \pm 0.02$ & $3.96 \pm 0.19$ & $0.40 \pm 0.01$ \\
 & Prediction \& Uncertainty & $0.79 \pm 0.03$ & $0.08 \pm 0.01$ & $0.09 \pm 0.02$ & $0.01 \pm 0.01$ & $0.02 \pm 0.03$ & $3.86 \pm 0.29$ & $0.40 \pm 0.02$ \\
 & CMARL                   & $0.81 \pm 0.03$ & $0.07 \pm 0.02$ & $0.09 \pm 0.02$ & $0.01 \pm 0.01$ & $\mathbf{0.01 \pm 0.02}$ & $3.37 \pm 0.16$ & $0.42 \pm 0.02$ \\
\midrule
\multirow{3}{*}{Early Fusion}
 & Prediction              & $0.77 \pm 0.02$ & $0.11 \pm 0.01$ & $0.06 \pm 0.02$ & $0.01 \pm 0.01$ & $0.05 \pm 0.02$ & $2.68 \pm 0.36$ & $0.43 \pm 0.01$ \\
 & Prediction \& Uncertainty & $0.79 \pm 0.02$ & $0.10 \pm 0.02 $ & $0.05 \pm 0.02$ & $0.01 \pm 0.02 $ & $0.05 \pm 0.01$ & $2.56 \pm 0.36$ & $0.43 \pm 0.01$ \\
 & CMARL                   & $0.82 \pm 0.02$ & $0.07 \pm 0.01$ & $0.05 \pm 0.01$ & $\mathbf{0.00 \pm 0.01}$ & $0.06 \pm 0.02$ & $2.37 \pm 0.19$ & $0.43 \pm 0.01$ \\
\midrule
\multirow{3}{*}{Middle Fusion}
 & Prediction              & $0.84 \pm 0.02$ & $0.06 \pm 0.01$ & $0.05 \pm 0.01$ & $0.01 \pm 0.01$ & $0.04 \pm 0.02$ & $2.17 \pm 0.26$ & $0.42 \pm 0.01$ \\
 & Prediction \& Uncertainty & $0.86 \pm 0.02$ & $0.05 \pm 0.01$ & $0.04 \pm 0.01$ & $0.01 \pm 0.01$ & $0.04 \pm 0.02$ & $2.20 \pm 0.16$ & $0.43 \pm 0.01$ \\
 & CMARL                   & $\mathbf{0.90 \pm 0.03}$ & $\mathbf{0.03 \pm 0.01}$ & $\mathbf{0.03 \pm 0.01}$ & $\mathbf{0.00 \pm 0.00}$ & $0.03 \pm 0.01$ & $\mathbf{2.07 \pm 0.19}$ & $\mathbf{0.45 \pm 0.03}$ \\
\bottomrule
\end{tabular}}
\end{table}

\subsection{Cost Limit}
\label{appendix:cost_abalation}

Table~\ref{tab:cost_limit_ablation} shows the trade-off controlled by the cost limit $\epsilon$. A larger cost limit allows more efficient behavior, resulting in shorter navigation time and path length, but also higher collision rates and lower social distance. Reducing $\epsilon$ encourages safer navigation, with $\epsilon=0.10$ achieving the best SR, joint-lowest human CR, lowest ITR, and highest SD. Although it slightly increases NT and PL, this setting provides the best overall balance between safety and efficiency.

\begin{table}[H]
\centering
\scriptsize
\caption{Ablation study on the cost limit $\epsilon$ in MAPPO-Lagrangian. A
smaller cost limit encourages socially compliant behavior, while a higher cost
limit allows more efficient goal-reaching behavior.}
\label{tab:cost_limit_ablation}
\setlength{\tabcolsep}{3pt}
\renewcommand{\arraystretch}{0.9}
\resizebox{1.0\textwidth}{!}{\begin{tabular}{lcccccccccc}
\toprule
\textbf{Cost Limit $\epsilon$} & \textbf{SR} $\uparrow$ & \textbf{Obstacle CR} $\downarrow$ & \textbf{Human CR} $\downarrow$ & \textbf{Robots CR} $\downarrow$ & \textbf{TR} $\downarrow$ & \textbf{NT} $\downarrow$ & \textbf{PL} $\downarrow$ & \textbf{AA} $\downarrow$ & \textbf{ITR} (\%) $\downarrow$ & \textbf{SD} $\uparrow$ \\
\midrule
$0.20$ & $0.87 \pm 0.03$ & $0.05 \pm 0.02$ & $0.04 \pm 0.01$ & $0.01 \pm 0.01$ & $\mathbf{0.02 \pm 0.01}$ & $\mathbf{9.37 \pm 0.35}$ & $\mathbf{11.35 \pm 0.17}$ & $0.13 \pm 0.03$ & $2.19 \pm 0.31$ & $0.43 \pm 0.03$ \\
$0.15$ & $0.89 \pm 0.03$ & $0.04 \pm 0.01$ & $\mathbf{0.03 \pm 0.01}$ & $\mathbf{0.00 \pm 0.01}$ & $\mathbf{0.02 \pm 0.01}$ & $9.46 \pm 0.26$ & $11.56 \pm 0.19$ & $0.13 \pm 0.04$ & $2.15 \pm 0.23$ & $0.43 \pm 0.03$ \\
$0.10$ & $\mathbf{0.90 \pm 0.03}$ & $\mathbf{0.03 \pm 0.01}$ & $\mathbf{0.03 \pm 0.01}$ & $\mathbf{0.00 \pm 0.00}$ & $0.03 \pm 0.01$ & $9.80 \pm 0.15$ & $11.98 \pm 0.29$ & $0.13 \pm 0.05$ & $\mathbf{2.07 \pm 0.19}$ & $\mathbf{0.45 \pm 0.03}$ \\
\bottomrule
\end{tabular}}
\end{table}

\section{Experimental Setup}
\label{appendix:experimental_setup}

\subsection{Environment}
\label{appendix:environment}

\begin{wrapfigure}{R}{0.5\linewidth}
\centering
\vspace{-0.3cm}
\includegraphics[width=\linewidth]{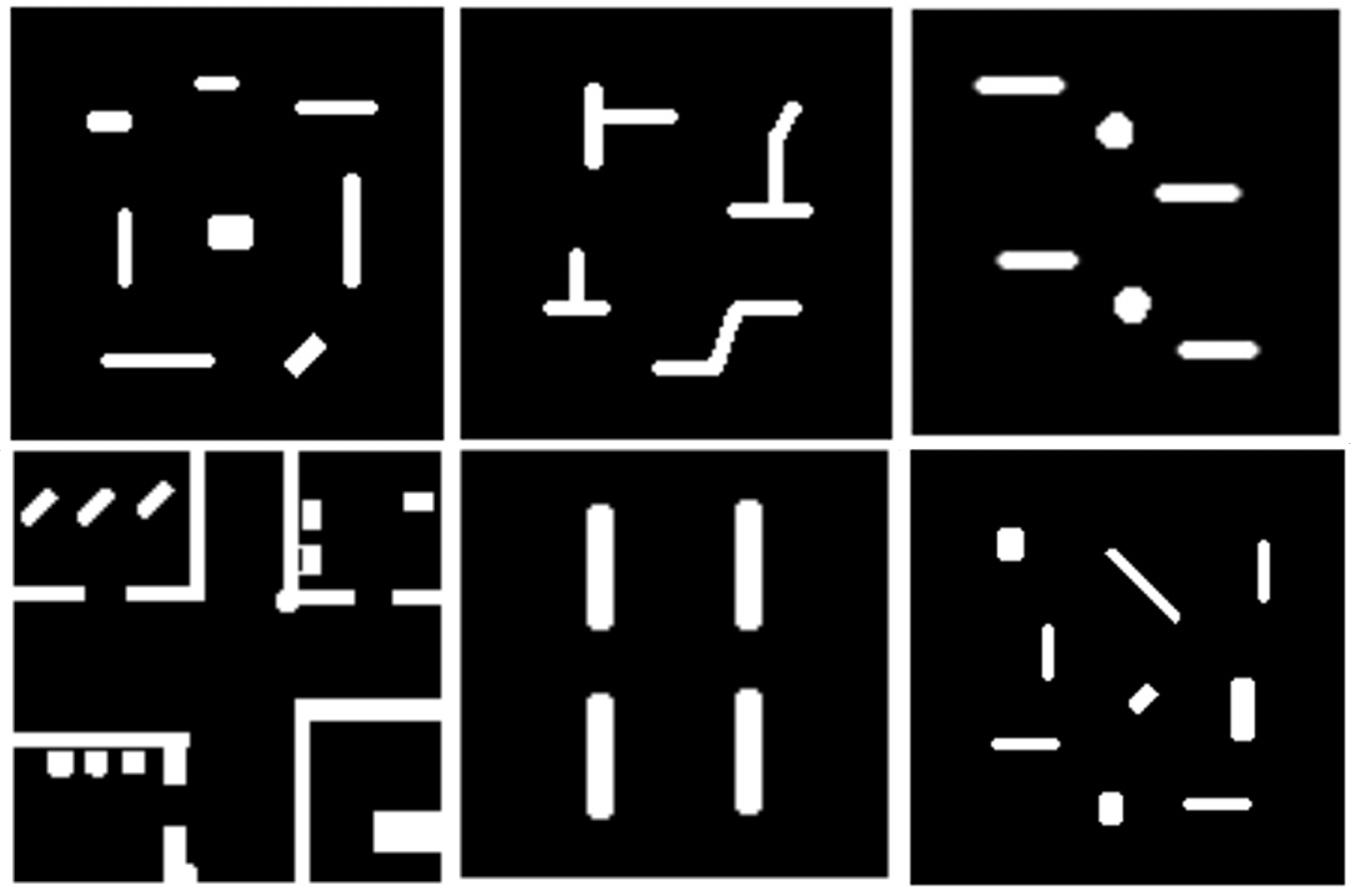}
\vspace{-0.3cm}
\caption{BEV Layouts.}
\vspace{-0.3cm}
\label{fig:simulation_layouts}
\end{wrapfigure}

The Unity3D simulator supports multiple robots, dynamic human agents, and diverse static obstacles. Static environments are generated by randomly placing convex polygons to represent non-navigable regions. Human agents navigate using Unity's built-in NavMesh system, which distinguishes walkable and non-walkable areas according to the obstacle layout. This enables pedestrians to move realistically in cluttered environments.
Pedestrian maximum speeds are randomized between \SI{0.8}{m/s} and \SI{1.5}{m/s} during regular simulation, and their radii are sampled between \SI{0.3}{m} and \SI{0.4}{m}. Each pedestrian randomly changes its goal every 3 to 5 seconds, increasing the stochasticity of crowd motion. Robots are modeled as invisible to humans following \cite{liu2023intention}, so robots are responsible for avoiding collisions. Each robot has a simulated 2D LiDAR sensor with a sensing range of \SI{7}{m}, radius \SI{0.2}{m}, and maximum speed \SI{1.0}{m/s}.

During policy training, three robots are initialized at randomized positions and assigned randomized goals in arenas populated with 5--10 pedestrians. As shown in Figure~\ref{fig:simulation_layouts}, the training environments consist of three layouts with different static obstacle configurations, with sizes ranging from $\SI{10}{m}\times\SI{10}{m}$ to $\SI{15}{m}\times\SI{15}{m}$. Three additional layouts are reserved for OOD evaluation. The in-distribution setting uses the same three layouts as training with 7--10 pedestrians. The OOD settings include unseen layouts, denser crowds with 12--15 pedestrians, and faster pedestrian motion with maximum speeds up to \SI{2}{m/s}.

For the offline forecasting dataset, LiDAR scans and robot poses are collected at \SI{5}{Hz}. Occupancy grids are generated with a resolution of \SI{0.1}{m} per cell and size $64 \times 64$. The history window is set to $\tau=5$. The dataset is collected from scenarios with three robots and 5--10 pedestrians across three environments, resulting in 361,152 training samples and 30,498 testing samples.

\subsection{Evaluation Metrics}
\label{appendix:evaluation_metrics}

We evaluate CoopUQ from two perspectives: occupancy forecasting quality and downstream navigation performance. For navigation, SR, CR, TR, and ITR use all evaluated robot episodes, whereas NT and PL use only successful episodes. AA averages episode-level values over episodes with at least three time steps. SD pools danger steps across all episodes.

\textbf{Structural Similarity Index Measure (SSIM).}
Let $\mathbf{m}_{t+\ell}^{i}$ denote the predicted occupancy grid map (OGM) of robot $i$ for future step $t+\ell$, and let $\hat{\mathbf{m}}_{t+\ell}^{i}$ denote the corresponding ground-truth OGM. SSIM evaluates the structural consistency between the predicted and ground-truth maps:
\begin{equation}
\mathrm{SSIM}(\mathbf{m}_{t+\ell}^{i}, \hat{\mathbf{m}}_{t+\ell}^{i})
=
\frac{
(2\mu_{\mathbf{m}}\mu_{\hat{\mathbf{m}}} + c_1)
(2\sigma_{\mathbf{m}\hat{\mathbf{m}}} + c_2)
}{
(\mu_{\mathbf{m}}^2 + \mu_{\hat{\mathbf{m}}}^2 + c_1)
(\sigma_{\mathbf{m}}^2 + \sigma_{\hat{\mathbf{m}}}^2 + c_2)
},
\end{equation}
where $\mu_{\mathbf{m}}$ and $\mu_{\hat{\mathbf{m}}}$ denote the mean occupancy values, $\sigma_{\mathbf{m}}^2$ and $\sigma_{\hat{\mathbf{m}}}^2$ denote the variances, and $\sigma_{\mathbf{m}\hat{\mathbf{m}}}$ denotes the covariance between the predicted and ground-truth. The constants $c_1$ and $c_2$ are small positive stabilizing terms. SSIM emphasizes spatial structure and obstacle geometry, making it suitable for evaluating whether the predicted OGM preserves the layout of occupied and free regions.

\textbf{Weighted Mean Squared Error (WMSE).}
WMSE measures the weighted pixel-wise squared error between the predicted OGM $\mathbf{m}_{t+\ell}^{i}$ and the ground-truth OGM $\hat{\mathbf{m}}_{t+\ell}^{i}$:
\begin{equation}
\mathrm{WMSE}(\mathbf{m}_{t+\ell}^{i}, \hat{\mathbf{m}}_{t+\ell}^{i})
=
\frac{
\sum_{g=1}^{|\Omega|} w_g
\left(m_{t+\ell,g}^{i} - \hat{m}_{t+\ell,g}^{i}\right)^2
}{
\sum_{g=1}^{|\Omega|} w_g
},
\end{equation}
where $\Omega$ denotes the set of grid cells, $m_{t+\ell,g}^{i}$ and $\hat{m}_{t+\ell,g}^{i}$ are the predicted and ground-truth occupancy values at grid cell $g$, respectively, and $w_g$ is the weight assigned to cell $g$. We compute $w_g$ using median frequency balancing to compensate for the class imbalance between occupied and free cells. Since occupied cells are usually sparse in OGMs, WMSE penalizes errors in occupied regions more heavily than standard MSE, making it more suitable for evaluating safety-critical occupancy prediction accuracy.

\textbf{Success Rate (SR).}
SR measures the ratio of successful robot episodes to all evaluated robot episodes. Let $\mathcal{E}$ denote the set of all evaluated robot episodes. Each episode contributes one terminal outcome: success, collision with a static obstacle, human, or robot, or timeout. An episode is considered successful if the robot reaches its goal within the time limit $T_{\max}=40\,\mathrm{s}$ without collision.

\textbf{Collision Rate (CR).}
Each CR is the fraction of all evaluated robot episodes terminated by a collision of the corresponding type: static obstacle, human, or robot.

\textbf{Timeout Rate (TR).}
TR is the fraction of all evaluated robot episodes that reach the time limit without success or collision.

\textbf{Navigation Time (NT).}
NT measures the average time required for the robot to reach the goal, computed only over successful episodes.

\textbf{Path Length (PL).}
Let $\mathcal{E}_{\mathrm{succ}}\subseteq\mathcal{E}$ denote the successful robot episodes. PL measures their average traveled distance:
\begin{equation}
\mathrm{PL}
=
\frac{1}{|\mathcal{E}_{\mathrm{succ}}|}
\sum_{e \in \mathcal{E}_{\mathrm{succ}}}
\sum_{t=1}^{L_e}
\left\|
\mathbf{p}_{e,t}^{r} - \mathbf{p}_{e,t-1}^{r}
\right\|_2,
\end{equation}
where $\mathbf{p}_{e,t}^{r}$ denotes the robot position at time step $t$ in episode $e$, and $L_e$ denotes the episode length in time steps.

\textbf{Average Acceleration (AA).}
AA measures trajectory smoothness. We compute a mean acceleration for each episode with at least three time steps, then average these episode-level values with equal weight.

\textbf{Intrusion Time Ratio (ITR).}
ITR measures how often the robot enters regions occupied by the ground-truth future positions of humans. Let $\mathcal{D}_{e,t}^{\mathrm{danger}}$ indicate whether the robot intrudes into a ground-truth future human-occupied region at time step $t$ in episode $e$. ITR is defined as
\begin{equation}
\mathrm{ITR}
=
\frac{1}{|\mathcal{E}|}
\sum_{e \in \mathcal{E}}
\left(
\frac{
\sum_{t=1}^{L_e}
\mathcal{D}_{e,t}^{\mathrm{danger}}
}{
L_e
}
\right)
\times 100\%.
\end{equation}
A lower ITR indicates that the robot more effectively avoids high-risk regions induced by future human motion.

\textbf{Social Distance (SD).}
SD measures the average minimum distance between the robot and nearby humans at danger steps. Let $\mathcal{T}_{e}^{\mathrm{danger}}$ denote the set of time steps in episode $e$ where a danger event occurs. For each danger step, we compute the minimum distance between the robot and all humans:
\begin{equation}
d_{e,t}^{\min}
=
\min_{j \in \mathcal{H}_{e,t}}
\left\|
\mathbf{p}_{e,t}^{r} - \mathbf{p}_{e,t}^{h_j}
\right\|_2,
\end{equation}
where $\mathbf{p}_{e,t}^{r}$ is the robot position, $\mathbf{p}_{e,t}^{h_j}$ is the position of human $j$, and $\mathcal{H}_{e,t}$ denotes the set of humans at time step $t$. The final SD is computed as
\begin{equation}
\mathrm{SD}
=
\frac{
\sum_{e \in \mathcal{E}}
\sum_{t \in \mathcal{T}_{e}^{\mathrm{danger}}}
d_{e,t}^{\min}
}{
\sum_{e \in \mathcal{E}}
|\mathcal{T}_{e}^{\mathrm{danger}}|
}.
\end{equation}
A higher SD indicates that the robot maintains a larger safety margin from humans when entering potentially risky regions.

\section{Implementation Details}
\label{appendix:implementation_details}

For cooperative navigation, we train a recurrent MAPPO policy with a Lagrangian cost constraint using three robots per simulation environment. The actor and critics use a three-layer MLP with 64 hidden units per layer, followed by a one-layer GRU. All networks are optimized by Adam with a learning rate of $5 \times 10^{-4}$, linearly decayed to zero. We use discount factor $\gamma=0.99$ and GAE parameter $\lambda=0.95$. Each PPO update uses 5 epochs, 2 mini-batches, and a recurrent data chunk length of 20. The PPO clipping coefficient is 0.2 and the entropy coefficient is 0.01. The value loss coefficient is 0.5, the Huber loss threshold is 10, and the maximum gradient norm is 0.5. We train with 8 parallel rollout threads for $10^6$ steps.

\subsection{Constrained Policy}
\label{appendix:constrained_policy}

The cost limit is set to 0.1, with additional ablations using limits of 0.15 and 0.2. The policy advantage is modified as
\begin{equation}
A_{\mathrm{mix}}
=
\frac{A - \eta A_{\mathrm{cost}}}{1+\eta},
\end{equation}
where $A$ is the reward advantage, $A_{\mathrm{cost}}$ is the cost advantage, and $\eta$ is the Lagrange multiplier. The multiplier learning rate is 0.003, the initial value is 0.1, and $\eta$ is clipped to $[0,5]$.
The safety cost is computed from the predicted occupancy and normalized
uncertainty within a local comfort region of radius $r_{\mathrm{safe}}=5$ grid cells, used for both $\mathcal{B}_{\text{occ}}^i$ and $\mathcal{B}_{\text{unc}}^i$. We set the occupancy threshold $\phi_{\text{occ}}=0.5$, the normalized uncertainty
threshold $\phi_{\text{unc}}=0.3$, and the weighting factors
$\lambda_{\text{occ}}=0.5$ and $\lambda_{\text{unc}}=0.5$. Each violating
cell incurs a distance-based penalty $w_d(x,y)=\max(0,\,r_{\mathrm{safe}}-d(x,y))$, so that threats closer to the ego robot are penalized more heavily. The resulting cost is scaled by the control interval $\Delta t=0.2\,\mathrm{s}$. This encourages the policy to avoid regions that are likely occupied or unreliable under stochastic forecasting.

\subsection{Stochastic OGM Forecasting}
\label{appendix:ogm_forecasting}

All forecasting models across fusion strategies use an OGM size of $64 \times 64$ with a resolution of \SI{0.1}{m} per cell, a history length $\tau=5$, and a prediction horizon $h=5$. LiDAR and pose sequences are sampled at \SI{5}{Hz}, so the horizon covers the next one second. We draw $M=8$ stochastic samples from the decoder. 
The cooperative OGM forecasting is trained with a reconstruction term and a KL
regularization term:
\begin{equation}
\mathcal{L}_{\mathrm{VAE}}
=
\mathcal{L}_{\mathrm{rec}}
+
\beta
\mathcal{L}_{\mathrm{KL}},
\end{equation}
where $\mathcal{L}_{\mathrm{rec}}$ measures the difference between the predicted
future OGMs and the ground-truth OGMs, and $\mathcal{L}_{\mathrm{KL}}$ regularizes
the learned latent distribution toward a standard Gaussian prior. We use a binary
occupancy reconstruction loss:
\begin{equation}
\mathcal{L}_{\mathrm{rec}}
=
-
\frac{1}{h|\Omega|}
\sum_{\ell=1}^{h}
\sum_{g \in \Omega}
\left[
\hat{m}_{t+\ell,g}^{i}
\log m_{t+\ell,g}^{i}
+
(1-\hat{m}_{t+\ell,g}^{i})
\log(1-m_{t+\ell,g}^{i})
\right],
\end{equation}
where $m_{t+\ell,g}^{i}$ and $\hat{m}_{t+\ell,g}^{i}$ denote the predicted and
ground-truth occupancy values at grid cell $g$, respectively. The KL term is
defined as
\begin{equation}
\mathcal{L}_{\mathrm{KL}}
=
D_{\mathrm{KL}}
\left(
q_{\phi}(\mathbf{z}_{t+1:t+h}^{i} \mid \mathbf{F}_{\mathrm{fuse}}^i)
\;\|\;
\mathcal{N}(\mathbf{0}, \mathbf{I})
\right).
\end{equation}
The coefficient $\beta$ balances reconstruction accuracy and latent-space
regularization. We set $\beta=0.01$. The model is trained for 20 epochs using
Adam with a learning rate of $10^{-4}$ and weight decay $10^{-3}$.

\end{document}